\documentclass{easychair}

\usepackage{float}
\usepackage{perpage}
\MakeSorted{figure}
\MakeSorted{table}

\usepackage{makeidx}

\newcommand{\xs}[1]{Section~\ref{#1}}

\newcommand{\tool}[1]{\texttt{#1}\index{Tools!#1}}

\newcommand{\api}[1]{\texttt{#1}\index{API!#1}}

\newcommand{\marf}[0]{MARF\index{MARF}\index{Frameworks!MARF}\index{Libraries!MARF}}

\newcommand{\lucidL}[1]{{$\mathit{Lucid}$}($L$) }

		{}

\def\myvert{\raise 2.27pt \hbox{\vrule depth 0pt height 8pt width 0.2mm}}
\def\myarrow{\hspace*{0.43mm}%
             \raise 2.29pt\hbox{\vrule depth 0pt height 8pt width 0.16mm}%
             \hspace*{-0.32mm}%
             $\longrightarrow$
             \ %
             }

\makeindex

\begin{document}

\author{Serguei A. Mokhov\\
Concordia University\\
Montreal, QC, Canada\\
\url{mokhov@cse.concordia.ca}}
\authorrunning{S. A. Mokhov}

\title{Complete Complementary Results Report of the MARF's NLP Approach to the DEFT 2010 Competition}
\titlerunning{Complete Result Set of the MARF Approach to DEFT'10}

\maketitle

\begin{abstract}
This companion paper complements the main DEFT'10 article~\cite{marf-deft} describing the MARF approach
to the DEFT'10 NLP challenge. This paper is aimed to present the complete result sets
of all the conducted experiments and their settings in the resulting tables
highlighting the approach and the best results, but also showing the worse
and the worst and their subsequent analysis.
This particular work focuses on application
of the {\marf}'s classical and NLP pipelines to identification tasks within various
francophone corpora to identify decades when certain articles
were published for the first track (Piste 1) and place of origin of
a publication (Piste 2), such as the journal and location (France vs. Quebec).
This is the forth iteration of the release of the results.
\end{abstract}

\tableofcontents
\listoftables

\section{Introduction}

The main article presenting the condensed summary as well
as methodology is in~\cite{marf-deft} (to be publicly available
in July 2010). We briefly cite the related work that is
founding to our approach.

\subsection{{\marf}}

The approach is based on the {\marf}'s framework implementation
documented in a number of related works~\cite{%
marf,%
marf02,%
marf-ai08,%
marf-c3s2e08,%
marf-file-type,%
marf-writer-ident,%
cryptolysis-sera2010,%
marf-nlp-framework,%
marf-leverage-sim-hsc09%
}

{\marf} is a general pattern recognition framework combined with
the corresponding implementation for a variety of audio, voice,
NLP, forensic, biometric authentication, and other pattern
recognition tasks.


\subsection{DEFT2010}

The DEFT2010 challenge description is in \cite{deft2010}.
It has two tracks--Piste~1: identifying the decade of francophone
publications (journal article fragments), and Piste 2: identifying
where the publication took place geographically--France or Quebec.
The DEFT2010 corpora were compiled by Cyril Grouin from a variety
of sources, namely Gallica, CEDROM-SNi, CNRTL, and ELDA with
restrictions on the CEDROM-SNi and ELDA data \cite{grouin-gallica-deft2010,%
le-devoir-cedrom-sni-deft2010,%
estrepublicain-cnrtl-deft2010,%
le-monde-elda-deft2010,%
la-presse-cedrom-sni-deft2010}.

\section{Methodology}
\label{sect:methodology}

The below is a partial list of the methodologies used in the
evaluation. It is to be updated with the next revisions of
the document; meanwhile the more complete list of the
techniques for the tasks, used, ongoing, and planned,
is in \cite{marf-deft}.

\subsection{Classical MARF Pipeline Approach}

The \api{DEFT2010App} application aggregated most or all of the
relevant features from the other tasks on speaker \api{SpeakerIdentApp}, language \api{LangIdentApp}, writer \api{WriterIdentApp}, file type,
identification and Zipf's Law \cite{marf-speaker-ident-app,%
marf-lang-ident-app,%
marf-zipf-law-app,%
marf-writer-ident-app%
}. Which combines the spectral text analysis \cite{spectral-text-analysis-taln2006}
with the traditional statistical NLP.

\subsection{Small Increase In the Article Text Size}

We collect titles as well as texts. For the training
purposes one way of experiment is to merge the two
effectively increasing a little bit the training
sample size for each article. The experiment is
to verify if adding titles to texts increases
the precision of detection.

\subsection{Titles vs. Texts}

For a 300-word article, its title may act as an abstract.
Often in the NLP, e.g. BioNLP, NLP techniques area
applied to the article abstracts rather than the complete
texts. A title in our sample can be considered as an
abstract on the same scale of a short article excerpt.
Thus we test the approach to see if titles alone are
sufficient to generate enough precision and increasing
the performance.

\section{The \api{DEFT2010App} {\marf} Application}
\label{sect:deft-app}
\index{DEFT2010App Application}

\subsection{Requirements}
\label{sect:deft-app-requirements}
\index{DEFT2010App Application!Requirements}

\begin{itemize}
	\item A current commodity computer running either Linux, MacOS X, or Windows with command line support via Cygwin \cite{cygwin}.
	\item JDK 1.5+
	\item (optional) To edit the code and text files, preferred Eclipse IDE \cite{eclipse}
	\item GNU \tool{make} \cite{gmake}
	\item \tool{tchs} 
	\item Perl 
\end{itemize}

\subsection{Options}
\label{sect:deft-app-options}
\index{DEFT2010App Application!Options}

In the below we briefly describe what each option means.
For the methodology and the scientific details, please
refer to the main article found in~\cite{marf-deft}.

\scriptsize
\begin{verbatim}
Usage:
  java DEFT2010App     --train <samples-dir> [options]        -- train mode
                       --single-train <sample> [options]      -- add a single sample to the training set
                       --ident <sample> [options]             -- identification mode
                       --batch-ident <samples-dir> [options]  -- batch identification mode
                       --train-nlp [ --debug ] [ OPTIONS ] <language> <corpus-file>
                       --ident-nlp [ --debug ] [ OPTIONS ] foo <bar|corpus-file>

                       --debug                                -- include verbose debug output
                       --gui                                  -- use GUI as a user interface
                       --stats=[per-config|per-writer|both]   -- display stats (default is per-config)
                       --best-score                           -- display best classification result
                       --reset                                -- reset stats
                       --version                              -- display version info
                       --help | -h                            -- display this help and exit

Options (one or more of the following):

Loaders:

  -wav          - assume WAVE files loading (default)
  -text         - assume loading of text samples
  -tiff         - assume loading of TIFF samples

Preprocessing:

  -silence      - remove silence (can be combined with any of the below)
  -noise        - remove noise (can be combined with any of the below)
  -raw          - no preprocessing
  -norm         - use just normalization, no filtering
  -low          - use low-pass FFT filter
  -high         - use high-pass FFT filter
  -boost        - use high-frequency-boost FFT preprocessor
  -band         - use band-pass FFT filter
  -bandstop     - use band-stop FFT filter
  -endp         - use endpointing
  -lowcfe       - use low-pass CFE filter
  -highcfe      - use high-pass CFE filter
  -bandcfe      - use band-pass CFE filter
  -bandstopcfe  - use band-stop CFE filter

Feature Extraction:

  -lpc          - use LPC
  -fft          - use FFT
  -minmax       - use Min/Max Amplitudes
  -randfe       - use random feature extraction
  -aggr         - use aggregated FFT+LPC feature extraction
  -f0           - use F0 (pitch, or fundamental frequency; NOT IMPLEMENTED
  -segm         - use Segmentation (NOT IMPLEMENTED)
  -cepstral     - use Cepstral analysis (NOT IMPLEMENTED)

Classification:

  -nn           - use Neural Network
  -cheb         - use Chebyshev Distance
  -eucl         - use Euclidean Distance
  -mink         - use Minkowski Distance
  -diff         - use Diff-Distance
  -zipf         - use Zipf's Law-based classifier
  -randcl       - use random classification
  -markov       - use Hidden Markov Models (NOT IMPLEMENTED)
  -hamming      - use Hamming Distance
  -cos          - use Cosine Similarity Measure

NLP/Ngrams/Smoothing:

  -interactive  - interactive mode for classification instead of reading from a file
  -char         - use characters as n-grams (should always be present for this app)

  -unigram      - use UNIGRAM model
  -bigram       - use BIGRAM model
  -trigram      - use TRIGRAM model

  -mle          - use MLE
  -add-one      - use Add-One smoothing
  -add-delta    - use Add-Delta (ELE, d=0.5) smoothing
  -witten-bell  - use Witten-Bell smoothing
  -good-turing  - use Good-Turing smoothing
  
Misc:

  -spectrogram  - dump spectrogram image after feature extraction
  -graph        - dump wave graph before preprocessing and after feature extraction
  <integer>     - expected subject ID
\end{verbatim}
\normalsize

\clearpage

\section{Results}
\label{sect:results}
\index{Results}

This is a large section consolidating the majority of the results.
Piste 1's results are in \xs{sect:results-piste1} and Piste 2's results
are in \xs{sect:results-piste2}.

\subsection{Piste 1 / Track 1: Decades}
\label{sect:results-piste1}
\index{Results!Piste 1}
\index{Results!Track 1}

Decades

\clearpage

\subsubsection{Testing on the evaluation data}

\begin{table}[H]
\caption{Consolidated results (piste1-result1-marf-eval-stats--fast), Part 1.}
\label{tab:piste1-result1-marf-eval-stats--fast-results1}
\small
\begin{minipage}[b]{\textwidth}
\centering
\begin{tabular}{|c|c|l|c|c|r|} \hline
Run \# & Guess & Configuration & GOOD & BAD & Precision,\%\\ \hline\hline
1 & 1st & -piste1 -ref -silence -bandstop -aggr -cos  & 331 & 2390 & 12.16\\ \hline
2 & 1st & -piste1 -ref -noise -raw -aggr -eucl  & 315 & 2406 & 11.58\\ \hline
3 & 1st & -piste1 -ref -silence -raw -aggr -eucl  & 315 & 2406 & 11.58\\ \hline
4 & 1st & -piste1 -ref -norm -fft -cos  & 315 & 2406 & 11.58\\ \hline
5 & 1st & -piste1 -ref -raw -aggr -eucl  & 315 & 2406 & 11.58\\ \hline
6 & 1st & -piste1 -ref -noise -raw -fft -cos  & 315 & 2406 & 11.58\\ \hline
7 & 1st & -piste1 -ref -silence -raw -fft -eucl  & 315 & 2406 & 11.58\\
\hline
\end{tabular}
\end{minipage}
\end{table}

\begin{table}[H]
\caption{Consolidated results (piste1-result1-marf-eval-stats--fast), Part 2.}
\label{tab:piste1-result1-marf-eval-stats--fast-results2}
\small
\begin{minipage}[b]{\textwidth}
\centering
\begin{tabular}{|c|c|l|c|c|r|} \hline
Run \# & Guess & Configuration & GOOD & BAD & Precision,\%\\ \hline\hline
50 & 2nd & -piste1 -ref -silence -bandstop -aggr -cos  & 555 & 2166 & 20.40\\ \hline
51 & 2nd & -piste1 -ref -noise -raw -aggr -eucl  & 572 & 2149 & 21.02\\ \hline
52 & 2nd & -piste1 -ref -silence -raw -aggr -eucl  & 572 & 2149 & 21.02\\ \hline
53 & 2nd & -piste1 -ref -norm -fft -cos  & 530 & 2191 & 19.48\\ \hline
54 & 2nd & -piste1 -ref -raw -aggr -eucl  & 572 & 2149 & 21.02\\ \hline
55 & 2nd & -piste1 -ref -noise -raw -fft -cos  & 532 & 2189 & 19.55\\ \hline
56 & 2nd & -piste1 -ref -silence -raw -fft -eucl  & 572 & 2149 & 21.02\\
\hline
\end{tabular}
\end{minipage}
\end{table}

\begin{table}[H]
\caption{Consolidated results (piste1-result1-marf-eval-stats--fast), Part 3.}
\label{tab:piste1-result1-marf-eval-stats--fast-results3}
\small
\begin{minipage}[b]{\textwidth}
\centering
\begin{tabular}{|c|c|l|c|c|r|} \hline
Run \# & Guess & Configuration & GOOD & BAD & Precision,\%\\ \hline\hline
1 & 1st & 1940 & 3852 & 5850 & 39.70\\ \hline
2 & 1st & 1830 & 2187 & 6094 & 26.41\\ \hline
3 & 1st & 1820 & 1571 & 6710 & 18.97\\ \hline
4 & 1st & 1810 & 1180 & 7101 & 14.25\\ \hline
5 & 1st & 1880 & 757 & 7524 & 9.14\\ \hline
6 & 1st & 1900 & 864 & 9083 & 8.69\\ \hline
7 & 1st & 1920 & 703 & 9097 & 7.17\\ \hline
8 & 1st & 1850 & 554 & 7727 & 6.69\\ \hline
9 & 1st & 1870 & 417 & 7864 & 5.04\\ \hline
10 & 1st & 1840 & 413 & 7868 & 4.99\\ \hline
11 & 1st & 1860 & 395 & 7935 & 4.74\\ \hline
12 & 1st & 1930 & 442 & 9358 & 4.51\\ \hline
13 & 1st & 1890 & 403 & 9250 & 4.17\\ \hline
14 & 1st & 1800 & 281 & 8000 & 3.39\\ \hline
15 & 1st & 1910 & 290 & 9559 & 2.94\\ \hline
16 & 2nd & 1940 & 4828 & 4874 & 49.76\\ \hline
17 & 2nd & 1830 & 3457 & 4824 & 41.75\\ \hline
18 & 2nd & 1820 & 3068 & 5213 & 37.05\\ \hline
19 & 2nd & 1810 & 1772 & 6509 & 21.40\\ \hline
20 & 2nd & 1880 & 1770 & 6511 & 21.37\\ \hline
21 & 2nd & 1900 & 1924 & 8023 & 19.34\\ \hline
22 & 2nd & 1920 & 1725 & 8075 & 17.60\\ \hline
23 & 2nd & 1850 & 1068 & 7213 & 12.90\\ \hline
24 & 2nd & 1870 & 881 & 7400 & 10.64\\ \hline
25 & 2nd & 1840 & 966 & 7315 & 11.67\\ \hline
26 & 2nd & 1860 & 884 & 7446 & 10.61\\ \hline
27 & 2nd & 1930 & 1150 & 8650 & 11.73\\ \hline
28 & 2nd & 1890 & 806 & 8847 & 8.35\\ \hline
29 & 2nd & 1800 & 978 & 7303 & 11.81\\ \hline
30 & 2nd & 1910 & 554 & 9295 & 5.62\\ \hline
\end{tabular}
\end{minipage}
\end{table}

\begin{table}[H]
\caption{Fichier \'{e}valu\'{e} : \tiny{\url{equipe_3_tache_1_execution_3-testing.sh-piste1-ref-noise-raw-aggr-cheb.prexml.xml}}}
\label{tab:equipe_3_tache_1_execution_3-testing.sh-piste1-ref-noise-raw-aggr-cheb.prexml.xml}
\small
\begin{minipage}[b]{\textwidth}
\centering
%
\end{minipage}
\end{table}
\begin{table}[H]
\caption{Fichier \'{e}valu\'{e} : \tiny{\url{equipe_3_tache_1_execution_3-testing.sh-piste1-ref-noise-raw-aggr-cos.prexml.xml}}}
\label{tab:equipe_3_tache_1_execution_3-testing.sh-piste1-ref-noise-raw-aggr-cos.prexml.xml}
\small
\begin{minipage}[b]{\textwidth}
\centering
%
\end{minipage}
\end{table}
\begin{table}[H]
\caption{Fichier \'{e}valu\'{e} : \tiny{\url{equipe_3_tache_1_execution_3-testing.sh-piste1-ref-noise-raw-aggr-diff.prexml.xml}}}
\label{tab:equipe_3_tache_1_execution_3-testing.sh-piste1-ref-noise-raw-aggr-diff.prexml.xml}
\small
\begin{minipage}[b]{\textwidth}
\centering
%
\end{minipage}
\end{table}
\begin{table}[H]
\caption{Fichier \'{e}valu\'{e} : \tiny{\url{equipe_3_tache_1_execution_3-testing.sh-piste1-ref-noise-raw-aggr-eucl.prexml.xml}}}
\label{tab:equipe_3_tache_1_execution_3-testing.sh-piste1-ref-noise-raw-aggr-eucl.prexml.xml}
\small
\begin{minipage}[b]{\textwidth}
\centering
%
\end{minipage}
\end{table}
\begin{table}[H]
\caption{Fichier \'{e}valu\'{e} : \tiny{\url{equipe_3_tache_1_execution_3-testing.sh-piste1-ref-noise-raw-fft-cheb.prexml.xml}}}
\label{tab:equipe_3_tache_1_execution_3-testing.sh-piste1-ref-noise-raw-fft-cheb.prexml.xml}
\small
\begin{minipage}[b]{\textwidth}
\centering
%
\end{minipage}
\end{table}
\begin{table}[H]
\caption{Fichier \'{e}valu\'{e} : \tiny{\url{equipe_3_tache_1_execution_3-testing.sh-piste1-ref-noise-raw-fft-cos.prexml.xml}}}
\label{tab:equipe_3_tache_1_execution_3-testing.sh-piste1-ref-noise-raw-fft-cos.prexml.xml}
\small
\begin{minipage}[b]{\textwidth}
\centering
%
\end{minipage}
\end{table}
\begin{table}[H]
\caption{Fichier \'{e}valu\'{e} : \tiny{\url{equipe_3_tache_1_execution_3-testing.sh-piste1-ref-noise-raw-fft-diff.prexml.xml}}}
\label{tab:equipe_3_tache_1_execution_3-testing.sh-piste1-ref-noise-raw-fft-diff.prexml.xml}
\small
\begin{minipage}[b]{\textwidth}
\centering
%
\end{minipage}
\end{table}
\begin{table}[H]
\caption{Fichier \'{e}valu\'{e} : \tiny{\url{equipe_3_tache_1_execution_3-testing.sh-piste1-ref-noise-raw-fft-eucl.prexml.xml}}}
\label{tab:equipe_3_tache_1_execution_3-testing.sh-piste1-ref-noise-raw-fft-eucl.prexml.xml}
\small
\begin{minipage}[b]{\textwidth}
\centering
%
\end{minipage}
\end{table}
\begin{table}[H]
\caption{Fichier \'{e}valu\'{e} : \tiny{\url{equipe_3_tache_1_execution_3-testing.sh-piste1-ref-norm-aggr-cheb.prexml.xml}}}
\label{tab:equipe_3_tache_1_execution_3-testing.sh-piste1-ref-norm-aggr-cheb.prexml.xml}
\small
\begin{minipage}[b]{\textwidth}
\centering
%
\end{minipage}
\end{table}
\begin{table}[H]
\caption{Fichier \'{e}valu\'{e} : \tiny{\url{equipe_3_tache_1_execution_3-testing.sh-piste1-ref-norm-aggr-cos.prexml.xml}}}
\label{tab:equipe_3_tache_1_execution_3-testing.sh-piste1-ref-norm-aggr-cos.prexml.xml}
\small
\begin{minipage}[b]{\textwidth}
\centering
%
\end{minipage}
\end{table}
\begin{table}[H]
\caption{Fichier \'{e}valu\'{e} : \tiny{\url{equipe_3_tache_1_execution_3-testing.sh-piste1-ref-norm-aggr-diff.prexml.xml}}}
\label{tab:equipe_3_tache_1_execution_3-testing.sh-piste1-ref-norm-aggr-diff.prexml.xml}
\small
\begin{minipage}[b]{\textwidth}
\centering
%
\end{minipage}
\end{table}
\begin{table}[H]
\caption{Fichier \'{e}valu\'{e} : \tiny{\url{equipe_3_tache_1_execution_3-testing.sh-piste1-ref-norm-aggr-eucl.prexml.xml}}}
\label{tab:equipe_3_tache_1_execution_3-testing.sh-piste1-ref-norm-aggr-eucl.prexml.xml}
\small
\begin{minipage}[b]{\textwidth}
\centering
%
\end{minipage}
\end{table}
\begin{table}[H]
\caption{Fichier \'{e}valu\'{e} : \tiny{\url{equipe_3_tache_1_execution_3-testing.sh-piste1-ref-norm-fft-cheb.prexml.xml}}}
\label{tab:equipe_3_tache_1_execution_3-testing.sh-piste1-ref-norm-fft-cheb.prexml.xml}
\small
\begin{minipage}[b]{\textwidth}
\centering
%
\end{minipage}
\end{table}
\begin{table}[H]
\caption{Fichier \'{e}valu\'{e} : \tiny{\url{equipe_3_tache_1_execution_3-testing.sh-piste1-ref-norm-fft-cos.prexml.xml}}}
\label{tab:equipe_3_tache_1_execution_3-testing.sh-piste1-ref-norm-fft-cos.prexml.xml}
\small
\begin{minipage}[b]{\textwidth}
\centering
%
\end{minipage}
\end{table}
\begin{table}[H]
\caption{Fichier \'{e}valu\'{e} : \tiny{\url{equipe_3_tache_1_execution_3-testing.sh-piste1-ref-norm-fft-diff.prexml.xml}}}
\label{tab:equipe_3_tache_1_execution_3-testing.sh-piste1-ref-norm-fft-diff.prexml.xml}
\small
\begin{minipage}[b]{\textwidth}
\centering
%
\end{minipage}
\end{table}
\begin{table}[H]
\caption{Fichier \'{e}valu\'{e} : \tiny{\url{equipe_3_tache_1_execution_3-testing.sh-piste1-ref-norm-fft-eucl.prexml.xml}}}
\label{tab:equipe_3_tache_1_execution_3-testing.sh-piste1-ref-norm-fft-eucl.prexml.xml}
\small
\begin{minipage}[b]{\textwidth}
\centering
%
\end{minipage}
\end{table}
\begin{table}[H]
\caption{Fichier \'{e}valu\'{e} : \tiny{\url{equipe_3_tache_1_execution_3-testing.sh-piste1-ref-raw-aggr-cheb.prexml.xml}}}
\label{tab:equipe_3_tache_1_execution_3-testing.sh-piste1-ref-raw-aggr-cheb.prexml.xml}
\small
\begin{minipage}[b]{\textwidth}
\centering
%
\end{minipage}
\end{table}
\begin{table}[H]
\caption{Fichier \'{e}valu\'{e} : \tiny{\url{equipe_3_tache_1_execution_3-testing.sh-piste1-ref-raw-aggr-cos.prexml.xml}}}
\label{tab:equipe_3_tache_1_execution_3-testing.sh-piste1-ref-raw-aggr-cos.prexml.xml}
\small
\begin{minipage}[b]{\textwidth}
\centering
%
\end{minipage}
\end{table}
\begin{table}[H]
\caption{Fichier \'{e}valu\'{e} : \tiny{\url{equipe_3_tache_1_execution_3-testing.sh-piste1-ref-raw-aggr-diff.prexml.xml}}}
\label{tab:equipe_3_tache_1_execution_3-testing.sh-piste1-ref-raw-aggr-diff.prexml.xml}
\small
\begin{minipage}[b]{\textwidth}
\centering
%
\end{minipage}
\end{table}
\begin{table}[H]
\caption{Fichier \'{e}valu\'{e} : \tiny{\url{equipe_3_tache_1_execution_3-testing.sh-piste1-ref-raw-aggr-eucl.prexml.xml}}}
\label{tab:equipe_3_tache_1_execution_3-testing.sh-piste1-ref-raw-aggr-eucl.prexml.xml}
\small
\begin{minipage}[b]{\textwidth}
\centering
%
\end{minipage}
\end{table}
\begin{table}[H]
\caption{Fichier \'{e}valu\'{e} : \tiny{\url{equipe_3_tache_1_execution_3-testing.sh-piste1-ref-raw-fft-cheb.prexml.xml}}}
\label{tab:equipe_3_tache_1_execution_3-testing.sh-piste1-ref-raw-fft-cheb.prexml.xml}
\small
\begin{minipage}[b]{\textwidth}
\centering
%
\end{minipage}
\end{table}
\begin{table}[H]
\caption{Fichier \'{e}valu\'{e} : \tiny{\url{equipe_3_tache_1_execution_3-testing.sh-piste1-ref-raw-fft-cos.prexml.xml}}}
\label{tab:equipe_3_tache_1_execution_3-testing.sh-piste1-ref-raw-fft-cos.prexml.xml}
\small
\begin{minipage}[b]{\textwidth}
\centering
%
\end{minipage}
\end{table}
\begin{table}[H]
\caption{Fichier \'{e}valu\'{e} : \tiny{\url{equipe_3_tache_1_execution_3-testing.sh-piste1-ref-raw-fft-diff.prexml.xml}}}
\label{tab:equipe_3_tache_1_execution_3-testing.sh-piste1-ref-raw-fft-diff.prexml.xml}
\small
\begin{minipage}[b]{\textwidth}
\centering
%
\end{minipage}
\end{table}
\begin{table}[H]
\caption{Fichier \'{e}valu\'{e} : \tiny{\url{equipe_3_tache_1_execution_3-testing.sh-piste1-ref-raw-fft-eucl.prexml.xml}}}
\label{tab:equipe_3_tache_1_execution_3-testing.sh-piste1-ref-raw-fft-eucl.prexml.xml}
\small
\begin{minipage}[b]{\textwidth}
\centering
%
\end{minipage}
\end{table}
\begin{table}[H]
\caption{Fichier \'{e}valu\'{e} : \tiny{\url{equipe_3_tache_1_execution_3-testing.sh-piste1-ref-silence-bandstop-aggr-cos.prexml.xml}}}
\label{tab:equipe_3_tache_1_execution_3-testing.sh-piste1-ref-silence-bandstop-aggr-cos.prexml.xml}
\small
\begin{minipage}[b]{\textwidth}
\centering
%
\end{minipage}
\end{table}
\begin{table}[H]
\caption{Fichier \'{e}valu\'{e} : \tiny{\url{equipe_3_tache_1_execution_3-testing.sh-piste1-ref-silence-noise-raw-aggr-cheb.prexml.xml}}}
\label{tab:equipe_3_tache_1_execution_3-testing.sh-piste1-ref-silence-noise-raw-aggr-cheb.prexml.xml}
\small
\begin{minipage}[b]{\textwidth}
\centering
%
\end{minipage}
\end{table}
\begin{table}[H]
\caption{Fichier \'{e}valu\'{e} : \tiny{\url{equipe_3_tache_1_execution_3-testing.sh-piste1-ref-silence-noise-raw-aggr-cos.prexml.xml}}}
\label{tab:equipe_3_tache_1_execution_3-testing.sh-piste1-ref-silence-noise-raw-aggr-cos.prexml.xml}
\small
\begin{minipage}[b]{\textwidth}
\centering
%
\end{minipage}
\end{table}
\begin{table}[H]
\caption{Fichier \'{e}valu\'{e} : \tiny{\url{equipe_3_tache_1_execution_3-testing.sh-piste1-ref-silence-noise-raw-aggr-diff.prexml.xml}}}
\label{tab:equipe_3_tache_1_execution_3-testing.sh-piste1-ref-silence-noise-raw-aggr-diff.prexml.xml}
\small
\begin{minipage}[b]{\textwidth}
\centering
%
\end{minipage}
\end{table}
\begin{table}[H]
\caption{Fichier \'{e}valu\'{e} : \tiny{\url{equipe_3_tache_1_execution_3-testing.sh-piste1-ref-silence-noise-raw-aggr-eucl.prexml.xml}}}
\label{tab:equipe_3_tache_1_execution_3-testing.sh-piste1-ref-silence-noise-raw-aggr-eucl.prexml.xml}
\small
\begin{minipage}[b]{\textwidth}
\centering
%
\end{minipage}
\end{table}
\begin{table}[H]
\caption{Fichier \'{e}valu\'{e} : \tiny{\url{equipe_3_tache_1_execution_3-testing.sh-piste1-ref-silence-noise-raw-fft-cheb.prexml.xml}}}
\label{tab:equipe_3_tache_1_execution_3-testing.sh-piste1-ref-silence-noise-raw-fft-cheb.prexml.xml}
\small
\begin{minipage}[b]{\textwidth}
\centering
%
\end{minipage}
\end{table}
\begin{table}[H]
\caption{Fichier \'{e}valu\'{e} : \tiny{\url{equipe_3_tache_1_execution_3-testing.sh-piste1-ref-silence-noise-raw-fft-cos.prexml.xml}}}
\label{tab:equipe_3_tache_1_execution_3-testing.sh-piste1-ref-silence-noise-raw-fft-cos.prexml.xml}
\small
\begin{minipage}[b]{\textwidth}
\centering
%
\end{minipage}
\end{table}
\begin{table}[H]
\caption{Fichier \'{e}valu\'{e} : \tiny{\url{equipe_3_tache_1_execution_3-testing.sh-piste1-ref-silence-noise-raw-fft-diff.prexml.xml}}}
\label{tab:equipe_3_tache_1_execution_3-testing.sh-piste1-ref-silence-noise-raw-fft-diff.prexml.xml}
\small
\begin{minipage}[b]{\textwidth}
\centering
%
\end{minipage}
\end{table}
\begin{table}[H]
\caption{Fichier \'{e}valu\'{e} : \tiny{\url{equipe_3_tache_1_execution_3-testing.sh-piste1-ref-silence-noise-raw-fft-eucl.prexml.xml}}}
\label{tab:equipe_3_tache_1_execution_3-testing.sh-piste1-ref-silence-noise-raw-fft-eucl.prexml.xml}
\small
\begin{minipage}[b]{\textwidth}
\centering
%
\end{minipage}
\end{table}
\begin{table}[H]
\caption{Fichier \'{e}valu\'{e} : \tiny{\url{equipe_3_tache_1_execution_3-testing.sh-piste1-ref-silence-norm-aggr-cheb.prexml.xml}}}
\label{tab:equipe_3_tache_1_execution_3-testing.sh-piste1-ref-silence-norm-aggr-cheb.prexml.xml}
\small
\begin{minipage}[b]{\textwidth}
\centering
%
\end{minipage}
\end{table}
\begin{table}[H]
\caption{Fichier \'{e}valu\'{e} : \tiny{\url{equipe_3_tache_1_execution_3-testing.sh-piste1-ref-silence-norm-aggr-cos.prexml.xml}}}
\label{tab:equipe_3_tache_1_execution_3-testing.sh-piste1-ref-silence-norm-aggr-cos.prexml.xml}
\small
\begin{minipage}[b]{\textwidth}
\centering
%
\end{minipage}
\end{table}
\begin{table}[H]
\caption{Fichier \'{e}valu\'{e} : \tiny{\url{equipe_3_tache_1_execution_3-testing.sh-piste1-ref-silence-norm-aggr-diff.prexml.xml}}}
\label{tab:equipe_3_tache_1_execution_3-testing.sh-piste1-ref-silence-norm-aggr-diff.prexml.xml}
\small
\begin{minipage}[b]{\textwidth}
\centering
%
\end{minipage}
\end{table}
\begin{table}[H]
\caption{Fichier \'{e}valu\'{e} : \tiny{\url{equipe_3_tache_1_execution_3-testing.sh-piste1-ref-silence-norm-aggr-eucl.prexml.xml}}}
\label{tab:equipe_3_tache_1_execution_3-testing.sh-piste1-ref-silence-norm-aggr-eucl.prexml.xml}
\small
\begin{minipage}[b]{\textwidth}
\centering
%
\end{minipage}
\end{table}
\begin{table}[H]
\caption{Fichier \'{e}valu\'{e} : \tiny{\url{equipe_3_tache_1_execution_3-testing.sh-piste1-ref-silence-norm-fft-cheb.prexml.xml}}}
\label{tab:equipe_3_tache_1_execution_3-testing.sh-piste1-ref-silence-norm-fft-cheb.prexml.xml}
\small
\begin{minipage}[b]{\textwidth}
\centering
%
\end{minipage}
\end{table}
\begin{table}[H]
\caption{Fichier \'{e}valu\'{e} : \tiny{\url{equipe_3_tache_1_execution_3-testing.sh-piste1-ref-silence-norm-fft-cos.prexml.xml}}}
\label{tab:equipe_3_tache_1_execution_3-testing.sh-piste1-ref-silence-norm-fft-cos.prexml.xml}
\small
\begin{minipage}[b]{\textwidth}
\centering
%
\end{minipage}
\end{table}
\begin{table}[H]
\caption{Fichier \'{e}valu\'{e} : \tiny{\url{equipe_3_tache_1_execution_3-testing.sh-piste1-ref-silence-norm-fft-diff.prexml.xml}}}
\label{tab:equipe_3_tache_1_execution_3-testing.sh-piste1-ref-silence-norm-fft-diff.prexml.xml}
\small
\begin{minipage}[b]{\textwidth}
\centering
%
\end{minipage}
\end{table}
\begin{table}[H]
\caption{Fichier \'{e}valu\'{e} : \tiny{\url{equipe_3_tache_1_execution_3-testing.sh-piste1-ref-silence-norm-fft-eucl.prexml.xml}}}
\label{tab:equipe_3_tache_1_execution_3-testing.sh-piste1-ref-silence-norm-fft-eucl.prexml.xml}
\small
\begin{minipage}[b]{\textwidth}
\centering
%
\end{minipage}
\end{table}
\begin{table}[H]
\caption{Fichier \'{e}valu\'{e} : \tiny{\url{equipe_3_tache_1_execution_3-testing.sh-piste1-ref-silence-raw-aggr-cheb.prexml.xml}}}
\label{tab:equipe_3_tache_1_execution_3-testing.sh-piste1-ref-silence-raw-aggr-cheb.prexml.xml}
\small
\begin{minipage}[b]{\textwidth}
\centering
%
\end{minipage}
\end{table}
\begin{table}[H]
\caption{Fichier \'{e}valu\'{e} : \tiny{\url{equipe_3_tache_1_execution_3-testing.sh-piste1-ref-silence-raw-aggr-cos.prexml.xml}}}
\label{tab:equipe_3_tache_1_execution_3-testing.sh-piste1-ref-silence-raw-aggr-cos.prexml.xml}
\small
\begin{minipage}[b]{\textwidth}
\centering
%
\end{minipage}
\end{table}
\begin{table}[H]
\caption{Fichier \'{e}valu\'{e} : \tiny{\url{equipe_3_tache_1_execution_3-testing.sh-piste1-ref-silence-raw-aggr-diff.prexml.xml}}}
\label{tab:equipe_3_tache_1_execution_3-testing.sh-piste1-ref-silence-raw-aggr-diff.prexml.xml}
\small
\begin{minipage}[b]{\textwidth}
\centering
%
\end{minipage}
\end{table}
\begin{table}[H]
\caption{Fichier \'{e}valu\'{e} : \tiny{\url{equipe_3_tache_1_execution_3-testing.sh-piste1-ref-silence-raw-aggr-eucl.prexml.xml}}}
\label{tab:equipe_3_tache_1_execution_3-testing.sh-piste1-ref-silence-raw-aggr-eucl.prexml.xml}
\small
\begin{minipage}[b]{\textwidth}
\centering
%
\end{minipage}
\end{table}
\begin{table}[H]
\caption{Fichier \'{e}valu\'{e} : \tiny{\url{equipe_3_tache_1_execution_3-testing.sh-piste1-ref-silence-raw-fft-cheb.prexml.xml}}}
\label{tab:equipe_3_tache_1_execution_3-testing.sh-piste1-ref-silence-raw-fft-cheb.prexml.xml}
\small
\begin{minipage}[b]{\textwidth}
\centering
%
\end{minipage}
\end{table}
\begin{table}[H]
\caption{Fichier \'{e}valu\'{e} : \tiny{\url{equipe_3_tache_1_execution_3-testing.sh-piste1-ref-silence-raw-fft-cos.prexml.xml}}}
\label{tab:equipe_3_tache_1_execution_3-testing.sh-piste1-ref-silence-raw-fft-cos.prexml.xml}
\small
\begin{minipage}[b]{\textwidth}
\centering
%
\end{minipage}
\end{table}
\begin{table}[H]
\caption{Fichier \'{e}valu\'{e} : \tiny{\url{equipe_3_tache_1_execution_3-testing.sh-piste1-ref-silence-raw-fft-diff.prexml.xml}}}
\label{tab:equipe_3_tache_1_execution_3-testing.sh-piste1-ref-silence-raw-fft-diff.prexml.xml}
\small
\begin{minipage}[b]{\textwidth}
\centering
%
\end{minipage}
\end{table}
\begin{table}[H]
\caption{Fichier \'{e}valu\'{e} : \tiny{\url{equipe_3_tache_1_execution_3-testing.sh-piste1-ref-silence-raw-fft-eucl.prexml.xml}}}
\label{tab:equipe_3_tache_1_execution_3-testing.sh-piste1-ref-silence-raw-fft-eucl.prexml.xml}
\small
\begin{minipage}[b]{\textwidth}
\centering
%
\end{minipage}
\end{table}

\subsubsection{Training and testing on training Data}

\begin{table}[H]
\caption{Consolidated results (piste1-result1-marf-stats), Part 1.}
\label{tab:piste1-result1-marf-stats-results1}
\small
\begin{minipage}[b]{\textwidth}
\centering
\begin{tabular}{|c|c|l|c|c|r|} \hline
Run \# & Guess & Configuration & GOOD & BAD & Precision,\%\\ \hline\hline
1 & 1st & -piste1 -norm -fft -cos  & 1025 & 2569 & 28.52\\ \hline
2 & 1st & -piste1 -silence -norm -aggr -cos  & 1025 & 2569 & 28.52\\ \hline
3 & 1st & -piste1 -silence -norm -fft -cos  & 1025 & 2569 & 28.52\\ \hline
4 & 1st & -piste1 -norm -aggr -cos  & 1025 & 2569 & 28.52\\ \hline
5 & 1st & -piste1 -raw -fft -cos  & 1023 & 2571 & 28.46\\ \hline
6 & 1st & -piste1 -silence -raw -aggr -cos  & 1023 & 2571 & 28.46\\ \hline
7 & 1st & -piste1 -silence -noise -raw -fft -cos  & 1023 & 2571 & 28.46\\
\hline
\end{tabular}
\end{minipage}
\end{table}

\begin{table}[H]
\caption{Consolidated results (piste1-result1-marf-stats), Part 2.}
\label{tab:piste1-result1-marf-stats-results2}
\small
\begin{minipage}[b]{\textwidth}
\centering
\begin{tabular}{|c|c|l|c|c|r|} \hline
Run \# & Guess & Configuration & GOOD & BAD & Precision,\%\\ \hline\hline
50 & 1st & -piste1 -silence -bandstop -fft -cos  & 860 & 2734 & 23.93\\ \hline
51 & 1st & -piste1 -bandstop -aggr -cos  & 855 & 2739 & 23.79\\ \hline
52 & 1st & -piste1 -bandstop -fft -cos  & 855 & 2739 & 23.79\\ \hline
53 & 1st & -piste1 -silence -high -aggr -cos  & 828 & 2766 & 23.04\\ \hline
54 & 1st & -piste1 -silence -high -fft -cos  & 826 & 2768 & 22.98\\ \hline
55 & 1st & -piste1 -silence -high -fft -eucl  & 797 & 2797 & 22.18\\ \hline
56 & 1st & -piste1 -silence -high -aggr -eucl  & 797 & 2797 & 22.18\\
\hline
\end{tabular}
\end{minipage}
\end{table}

\begin{table}[H]
\caption{Consolidated results (piste1-result1-marf-stats), Part 3.}
\label{tab:piste1-result1-marf-stats-results3}
\small
\begin{minipage}[b]{\textwidth}
\centering
\begin{tabular}{|c|c|l|c|c|r|} \hline
Run \# & Guess & Configuration & GOOD & BAD & Precision,\%\\ \hline\hline
100 & 1st & -piste1 -silence -noise -norm -aggr -diff  & 730 & 2864 & 20.31\\ \hline
101 & 1st & -piste1 -silence -noise -band -aggr -cos  & 730 & 2864 & 20.31\\ \hline
102 & 1st & -piste1 -silence -noise -norm -aggr -cheb  & 729 & 2865 & 20.28\\ \hline
103 & 1st & -piste1 -silence -low -aggr -cheb  & 729 & 2865 & 20.28\\ \hline
104 & 1st & -piste1 -silence -noise -band -fft -cos  & 728 & 2866 & 20.26\\ \hline
105 & 1st & -piste1 -low -fft -cheb  & 728 & 2866 & 20.26\\ \hline
106 & 1st & -piste1 -low -aggr -cheb  & 728 & 2866 & 20.26\\
\hline
\end{tabular}
\end{minipage}
\end{table}

\begin{table}[H]
\caption{Consolidated results (piste1-result1-marf-stats), Part 4.}
\label{tab:piste1-result1-marf-stats-results4}
\small
\begin{minipage}[b]{\textwidth}
\centering
\begin{tabular}{|c|c|l|c|c|r|} \hline
Run \# & Guess & Configuration & GOOD & BAD & Precision,\%\\ \hline\hline
150 & 1st & -piste1 -high -fft -cheb  & 640 & 2954 & 17.81\\ \hline
151 & 1st & -piste1 -silence -noise -endp -fft -cos  & 638 & 2956 & 17.75\\ \hline
152 & 1st & -piste1 -silence -noise -endp -aggr -cos  & 638 & 2956 & 17.75\\ \hline
153 & 1st & -piste1 -endp -aggr -diff  & 637 & 2957 & 17.72\\ \hline
154 & 1st & -piste1 -silence -endp -fft -diff  & 637 & 2957 & 17.72\\ \hline
155 & 1st & -piste1 -silence -endp -fft -cheb  & 637 & 2957 & 17.72\\ \hline
156 & 1st & -piste1 -endp -fft -diff  & 637 & 2957 & 17.72\\
\hline
\end{tabular}
\end{minipage}
\end{table}

\begin{table}[H]
\caption{Consolidated results (piste1-result1-marf-stats), Part 5.}
\label{tab:piste1-result1-marf-stats-results5}
\small
\begin{minipage}[b]{\textwidth}
\centering
\begin{tabular}{|c|c|l|c|c|r|} \hline
Run \# & Guess & Configuration & GOOD & BAD & Precision,\%\\ \hline\hline
200 & 1st & -piste1 -endp -fft -eucl  & 527 & 3067 & 14.66\\ \hline
201 & 1st & -piste1 -silence -endp -aggr -eucl  & 527 & 3067 & 14.66\\ \hline
202 & 1st & -piste1 -endp -aggr -eucl  & 527 & 3067 & 14.66\\ \hline
203 & 1st & -piste1 -silence -endp -fft -eucl  & 527 & 3067 & 14.66\\ \hline
204 & 1st & -piste1 -bandstop -fft -mink  & 526 & 3068 & 14.64\\ \hline
205 & 1st & -piste1 -noise -endp -fft -diff  & 526 & 3068 & 14.64\\ \hline
206 & 1st & -piste1 -bandstop -aggr -mink  & 526 & 3068 & 14.64\\
\hline
\end{tabular}
\end{minipage}
\end{table}

\begin{table}[H]
\caption{Consolidated results (piste1-result1-marf-stats), Part 6.}
\label{tab:piste1-result1-marf-stats-results6}
\small
\begin{minipage}[b]{\textwidth}
\centering
\begin{tabular}{|c|c|l|c|c|r|} \hline
Run \# & Guess & Configuration & GOOD & BAD & Precision,\%\\ \hline\hline
250 & 1st & -piste1 -noise -bandstop -fft -mink  & 423 & 3171 & 11.77\\ \hline
251 & 1st & -piste1 -noise -endp -fft -mink  & 423 & 3171 & 11.77\\ \hline
252 & 1st & -piste1 -silence -noise -endp -aggr -mink  & 419 & 3175 & 11.66\\ \hline
253 & 1st & -piste1 -silence -noise -endp -fft -mink  & 419 & 3175 & 11.66\\ \hline
254 & 1st & -piste1 -endp -fft -mink  & 417 & 3177 & 11.60\\ \hline
255 & 1st & -piste1 -silence -endp -aggr -mink  & 417 & 3177 & 11.60\\ \hline
256 & 1st & -piste1 -endp -aggr -mink  & 417 & 3177 & 11.60\\
\hline
\end{tabular}
\end{minipage}
\end{table}

\begin{table}[H]
\caption{Consolidated results (piste1-result1-marf-stats), Part 7.}
\label{tab:piste1-result1-marf-stats-results7}
\small
\begin{minipage}[b]{\textwidth}
\centering
\begin{tabular}{|c|c|l|c|c|r|} \hline
Run \# & Guess & Configuration & GOOD & BAD & Precision,\%\\ \hline\hline
300 & 1st & -piste1 -silence -endp -lpc -cheb  & 353 & 3241 & 9.82\\ \hline
301 & 1st & -piste1 -endp -lpc -cheb  & 353 & 3241 & 9.82\\ \hline
302 & 1st & -piste1 -silence -noise -endp -minmax -cheb  & 352 & 3242 & 9.79\\ \hline
303 & 1st & -piste1 -endp -minmax -cheb  & 351 & 3243 & 9.77\\ \hline
304 & 1st & -piste1 -silence -endp -minmax -cheb  & 351 & 3243 & 9.77\\ \hline
305 & 1st & -piste1 -noise -high -fft -cos  & 348 & 3246 & 9.68\\ \hline
306 & 1st & -piste1 -noise -low -minmax -eucl  & 347 & 3247 & 9.65\\
\hline
\end{tabular}
\end{minipage}
\end{table}

\begin{table}[H]
\caption{Consolidated results (piste1-result1-marf-stats), Part 8.}
\label{tab:piste1-result1-marf-stats-results8}
\small
\begin{minipage}[b]{\textwidth}
\centering
\begin{tabular}{|c|c|l|c|c|r|} \hline
Run \# & Guess & Configuration & GOOD & BAD & Precision,\%\\ \hline\hline
350 & 1st & -piste1 -noise -norm -minmax -cos  & 329 & 3265 & 9.15\\ \hline
351 & 1st & -piste1 -noise -norm -minmax -eucl  & 329 & 3265 & 9.15\\ \hline
352 & 1st & -piste1 -band -minmax -mink  & 329 & 3265 & 9.15\\ \hline
353 & 1st & -piste1 -silence -band -minmax -mink  & 329 & 3265 & 9.15\\ \hline
354 & 1st & -piste1 -low -minmax -cos  & 329 & 3265 & 9.15\\ \hline
355 & 1st & -piste1 -endp -minmax -eucl  & 329 & 3265 & 9.15\\ \hline
356 & 1st & -piste1 -silence -endp -minmax -eucl  & 329 & 3265 & 9.15\\
\hline
\end{tabular}
\end{minipage}
\end{table}

\begin{table}[H]
\caption{Consolidated results (piste1-result1-marf-stats), Part 9.}
\label{tab:piste1-result1-marf-stats-results9}
\small
\begin{minipage}[b]{\textwidth}
\centering
\begin{tabular}{|c|c|l|c|c|r|} \hline
Run \# & Guess & Configuration & GOOD & BAD & Precision,\%\\ \hline\hline
400 & 1st & -piste1 -bandstop -minmax -cheb  & 318 & 3276 & 8.85\\ \hline
401 & 1st & -piste1 -noise -band -minmax -eucl  & 317 & 3277 & 8.82\\ \hline
402 & 1st & -piste1 -silence -noise -band -minmax -eucl  & 317 & 3277 & 8.82\\ \hline
403 & 1st & -piste1 -high -minmax -eucl  & 315 & 3279 & 8.76\\ \hline
404 & 1st & -piste1 -silence -high -minmax -eucl  & 315 & 3279 & 8.76\\ \hline
405 & 1st & -piste1 -band -minmax -cheb  & 314 & 3280 & 8.74\\ \hline
406 & 1st & -piste1 -silence -band -minmax -cheb  & 314 & 3280 & 8.74\\
\hline
\end{tabular}
\end{minipage}
\end{table}

\begin{table}[H]
\caption{Consolidated results (piste1-result1-marf-stats), Part 10.}
\label{tab:piste1-result1-marf-stats-results10}
\small
\begin{minipage}[b]{\textwidth}
\centering
\begin{tabular}{|c|c|l|c|c|r|} \hline
Run \# & Guess & Configuration & GOOD & BAD & Precision,\%\\ \hline\hline
450 & 1st & -piste1 -high -minmax -diff  & 301 & 3293 & 8.38\\ \hline
451 & 1st & -piste1 -silence -high -minmax -diff  & 301 & 3293 & 8.38\\ \hline
452 & 1st & -piste1 -band -minmax -diff  & 301 & 3293 & 8.38\\ \hline
453 & 1st & -piste1 -silence -band -minmax -diff  & 301 & 3293 & 8.38\\ \hline
454 & 1st & -piste1 -silence -noise -bandstop -lpc -cos  & 301 & 3293 & 8.38\\ \hline
455 & 1st & -piste1 -noise -bandstop -aggr -hamming  & 301 & 3293 & 8.38\\ \hline
456 & 1st & -piste1 -high -minmax -mink  & 301 & 3293 & 8.38\\
\hline
\end{tabular}
\end{minipage}
\end{table}

\begin{table}[H]
\caption{Consolidated results (piste1-result1-marf-stats), Part 11.}
\label{tab:piste1-result1-marf-stats-results11}
\small
\begin{minipage}[b]{\textwidth}
\centering
\begin{tabular}{|c|c|l|c|c|r|} \hline
Run \# & Guess & Configuration & GOOD & BAD & Precision,\%\\ \hline\hline
500 & 1st & -piste1 -noise -norm -lpc -eucl  & 287 & 3307 & 7.99\\ \hline
501 & 1st & -piste1 -silence -noise -bandstop -minmax -diff  & 286 & 3308 & 7.96\\ \hline
502 & 1st & -piste1 -low -lpc -diff  & 286 & 3308 & 7.96\\ \hline
503 & 1st & -piste1 -silence -noise -bandstop -lpc -diff  & 286 & 3308 & 7.96\\ \hline
504 & 1st & -piste1 -noise -norm -lpc -diff  & 286 & 3308 & 7.96\\ \hline
505 & 1st & -piste1 -silence -noise -band -lpc -eucl  & 286 & 3308 & 7.96\\ \hline
506 & 1st & -piste1 -silence -high -lpc -cos  & 286 & 3308 & 7.96\\
\hline
\end{tabular}
\end{minipage}
\end{table}

\begin{table}[H]
\caption{Consolidated results (piste1-result1-marf-stats), Part 12.}
\label{tab:piste1-result1-marf-stats-results12}
\small
\begin{minipage}[b]{\textwidth}
\centering
\begin{tabular}{|c|c|l|c|c|r|} \hline
Run \# & Guess & Configuration & GOOD & BAD & Precision,\%\\ \hline\hline
550 & 1st & -piste1 -silence -raw -lpc -hamming  & 275 & 3319 & 7.65\\ \hline
551 & 1st & -piste1 -norm -lpc -hamming  & 275 & 3319 & 7.65\\ \hline
552 & 1st & -piste1 -silence -norm -lpc -hamming  & 275 & 3319 & 7.65\\ \hline
553 & 1st & -piste1 -bandstop -lpc -diff  & 275 & 3319 & 7.65\\ \hline
554 & 1st & -piste1 -noise -raw -lpc -hamming  & 275 & 3319 & 7.65\\ \hline
555 & 1st & -piste1 -noise -norm -lpc -mink  & 274 & 3320 & 7.62\\ \hline
556 & 1st & -piste1 -silence -noise -high -lpc -cheb  & 274 & 3320 & 7.62\\
\hline
\end{tabular}
\end{minipage}
\end{table}

\begin{table}[H]
\caption{Consolidated results (piste1-result1-marf-stats), Part 13.}
\label{tab:piste1-result1-marf-stats-results13}
\small
\begin{minipage}[b]{\textwidth}
\centering
\begin{tabular}{|c|c|l|c|c|r|} \hline
Run \# & Guess & Configuration & GOOD & BAD & Precision,\%\\ \hline\hline
600 & 1st & -piste1 -low -aggr -hamming  & 259 & 3335 & 7.21\\ \hline
601 & 1st & -piste1 -silence -noise -high -minmax -hamming  & 259 & 3335 & 7.21\\ \hline
602 & 1st & -piste1 -noise -high -minmax -hamming  & 259 & 3335 & 7.21\\ \hline
603 & 1st & -piste1 -silence -noise -low -aggr -hamming  & 258 & 3336 & 7.18\\ \hline
604 & 1st & -piste1 -silence -noise -low -fft -hamming  & 257 & 3337 & 7.15\\ \hline
605 & 1st & -piste1 -silence -high -lpc -cheb  & 257 & 3337 & 7.15\\ \hline
606 & 1st & -piste1 -silence -high -lpc -mink  & 257 & 3337 & 7.15\\
\hline
\end{tabular}
\end{minipage}
\end{table}

\begin{table}[H]
\caption{Consolidated results (piste1-result1-marf-stats), Part 14.}
\label{tab:piste1-result1-marf-stats-results14}
\small
\begin{minipage}[b]{\textwidth}
\centering
\begin{tabular}{|c|c|l|c|c|r|} \hline
Run \# & Guess & Configuration & GOOD & BAD & Precision,\%\\ \hline\hline
650 & 1st & -piste1 -noise -highpassboost -fft -cos  & 252 & 3342 & 7.01\\ \hline
651 & 1st & -piste1 -noise -boost -lpc -diff  & 252 & 3342 & 7.01\\ \hline
652 & 1st & -piste1 -noise -boost -aggr -hamming  & 252 & 3342 & 7.01\\ \hline
653 & 1st & -piste1 -noise -highpassboost -fft -mink  & 252 & 3342 & 7.01\\ \hline
654 & 1st & -piste1 -silence -highpassboost -aggr -eucl  & 252 & 3342 & 7.01\\ \hline
655 & 1st & -piste1 -silence -highpassboost -fft -eucl  & 252 & 3342 & 7.01\\ \hline
656 & 1st & -piste1 -noise -boost -lpc -cos  & 252 & 3342 & 7.01\\
\hline
\end{tabular}
\end{minipage}
\end{table}

\begin{table}[H]
\caption{Consolidated results (piste1-result1-marf-stats), Part 15.}
\label{tab:piste1-result1-marf-stats-results15}
\small
\begin{minipage}[b]{\textwidth}
\centering
\begin{tabular}{|c|c|l|c|c|r|} \hline
Run \# & Guess & Configuration & GOOD & BAD & Precision,\%\\ \hline\hline
700 & 1st & -piste1 -boost -lpc -diff  & 252 & 3342 & 7.01\\ \hline
701 & 1st & -piste1 -silence -noise -boost -lpc -hamming  & 252 & 3342 & 7.01\\ \hline
702 & 1st & -piste1 -highpassboost -minmax -cheb  & 252 & 3342 & 7.01\\ \hline
703 & 1st & -piste1 -highpassboost -minmax -hamming  & 252 & 3342 & 7.01\\ \hline
704 & 1st & -piste1 -silence -noise -boost -aggr -hamming  & 252 & 3342 & 7.01\\ \hline
705 & 1st & -piste1 -boost -aggr -eucl  & 252 & 3342 & 7.01\\ \hline
706 & 1st & -piste1 -noise -bandstop -lpc -hamming  & 252 & 3342 & 7.01\\
\hline
\end{tabular}
\end{minipage}
\end{table}

\begin{table}[H]
\caption{Consolidated results (piste1-result1-marf-stats), Part 16.}
\label{tab:piste1-result1-marf-stats-results16}
\small
\begin{minipage}[b]{\textwidth}
\centering
\begin{tabular}{|c|c|l|c|c|r|} \hline
Run \# & Guess & Configuration & GOOD & BAD & Precision,\%\\ \hline\hline
750 & 1st & -piste1 -silence -highpassboost -fft -hamming  & 252 & 3342 & 7.01\\ \hline
751 & 1st & -piste1 -noise -boost -minmax -cheb  & 252 & 3342 & 7.01\\ \hline
752 & 1st & -piste1 -highpassboost -minmax -eucl  & 252 & 3342 & 7.01\\ \hline
753 & 1st & -piste1 -silence -boost -lpc -hamming  & 252 & 3342 & 7.01\\ \hline
754 & 1st & -piste1 -boost -minmax -cheb  & 252 & 3342 & 7.01\\ \hline
755 & 1st & -piste1 -silence -highpassboost -aggr -diff  & 252 & 3342 & 7.01\\ \hline
756 & 1st & -piste1 -silence -highpassboost -fft -diff  & 252 & 3342 & 7.01\\
\hline
\end{tabular}
\end{minipage}
\end{table}

\begin{table}[H]
\caption{Consolidated results (piste1-result1-marf-stats), Part 17.}
\label{tab:piste1-result1-marf-stats-results17}
\small
\begin{minipage}[b]{\textwidth}
\centering
\begin{tabular}{|c|c|l|c|c|r|} \hline
Run \# & Guess & Configuration & GOOD & BAD & Precision,\%\\ \hline\hline
800 & 1st & -piste1 -noise -endp -lpc -hamming  & 245 & 3349 & 6.82\\ \hline
801 & 1st & -piste1 -noise -norm -lpc -hamming  & 245 & 3349 & 6.82\\ \hline
802 & 1st & -piste1 -low -lpc -hamming  & 245 & 3349 & 6.82\\ \hline
803 & 1st & -piste1 -silence -noise -high -minmax -mink  & 245 & 3349 & 6.82\\ \hline
804 & 1st & -piste1 -norm -fft -hamming  & 244 & 3350 & 6.79\\ \hline
805 & 1st & -piste1 -silence -norm -fft -hamming  & 244 & 3350 & 6.79\\ \hline
806 & 1st & -piste1 -silence -noise -band -fft -hamming  & 244 & 3350 & 6.79\\
\hline
\end{tabular}
\end{minipage}
\end{table}

\begin{table}[H]
\caption{Consolidated results (piste1-result1-marf-stats), Part 18.}
\label{tab:piste1-result1-marf-stats-results18}
\small
\begin{minipage}[b]{\textwidth}
\centering
\begin{tabular}{|c|c|l|c|c|r|} \hline
Run \# & Guess & Configuration & GOOD & BAD & Precision,\%\\ \hline\hline
850 & 2nd & -piste1 -norm -fft -cheb  & 1490 & 2104 & 41.46\\ \hline
851 & 2nd & -piste1 -norm -aggr -cheb  & 1489 & 2105 & 41.43\\ \hline
852 & 2nd & -piste1 -silence -norm -fft -cheb  & 1490 & 2104 & 41.46\\ \hline
853 & 2nd & -piste1 -silence -norm -fft -diff  & 1486 & 2108 & 41.35\\ \hline
854 & 2nd & -piste1 -norm -fft -diff  & 1486 & 2108 & 41.35\\ \hline
855 & 2nd & -piste1 -silence -noise -raw -fft -cheb  & 1488 & 2106 & 41.40\\ \hline
856 & 2nd & -piste1 -noise -raw -fft -cheb  & 1488 & 2106 & 41.40\\
\hline
\end{tabular}
\end{minipage}
\end{table}

\begin{table}[H]
\caption{Consolidated results (piste1-result1-marf-stats), Part 19.}
\label{tab:piste1-result1-marf-stats-results19}
\small
\begin{minipage}[b]{\textwidth}
\centering
\begin{tabular}{|c|c|l|c|c|r|} \hline
Run \# & Guess & Configuration & GOOD & BAD & Precision,\%\\ \hline\hline
900 & 2nd & -piste1 -low -fft -cos  & 1300 & 2294 & 36.17\\ \hline
901 & 2nd & -piste1 -noise -norm -aggr -cos  & 1301 & 2293 & 36.20\\ \hline
902 & 2nd & -piste1 -noise -norm -fft -cos  & 1300 & 2294 & 36.17\\ \hline
903 & 2nd & -piste1 -noise -low -aggr -cos  & 1294 & 2300 & 36.00\\ \hline
904 & 2nd & -piste1 -noise -low -fft -cos  & 1295 & 2299 & 36.03\\ \hline
905 & 2nd & -piste1 -bandstop -aggr -diff  & 1276 & 2318 & 35.50\\ \hline
906 & 2nd & -piste1 -bandstop -fft -diff  & 1274 & 2320 & 35.45\\
\hline
\end{tabular}
\end{minipage}
\end{table}

\begin{table}[H]
\caption{Consolidated results (piste1-result1-marf-stats), Part 20.}
\label{tab:piste1-result1-marf-stats-results20}
\small
\begin{minipage}[b]{\textwidth}
\centering
\begin{tabular}{|c|c|l|c|c|r|} \hline
Run \# & Guess & Configuration & GOOD & BAD & Precision,\%\\ \hline\hline
950 & 2nd & -piste1 -endp -fft -cos  & 1161 & 2433 & 32.30\\ \hline
951 & 2nd & -piste1 -endp -aggr -cos  & 1161 & 2433 & 32.30\\ \hline
952 & 2nd & -piste1 -silence -endp -fft -cos  & 1161 & 2433 & 32.30\\ \hline
953 & 2nd & -piste1 -silence -band -aggr -cos  & 1217 & 2377 & 33.86\\ \hline
954 & 2nd & -piste1 -noise -band -aggr -eucl  & 1195 & 2399 & 33.25\\ \hline
955 & 2nd & -piste1 -silence -band -fft -cos  & 1215 & 2379 & 33.81\\ \hline
956 & 2nd & -piste1 -noise -band -fft -eucl  & 1196 & 2398 & 33.28\\
\hline
\end{tabular}
\end{minipage}
\end{table}

\begin{table}[H]
\caption{Consolidated results (piste1-result1-marf-stats), Part 21.}
\label{tab:piste1-result1-marf-stats-results21}
\small
\begin{minipage}[b]{\textwidth}
\centering
\begin{tabular}{|c|c|l|c|c|r|} \hline
Run \# & Guess & Configuration & GOOD & BAD & Precision,\%\\ \hline\hline
1000 & 2nd & -piste1 -noise -low -fft -eucl  & 1096 & 2498 & 30.50\\ \hline
1001 & 2nd & -piste1 -band -fft -diff  & 1084 & 2510 & 30.16\\ \hline
1002 & 2nd & -piste1 -band -aggr -diff  & 1084 & 2510 & 30.16\\ \hline
1003 & 2nd & -piste1 -silence -noise -band -fft -diff  & 1041 & 2553 & 28.96\\ \hline
1004 & 2nd & -piste1 -silence -noise -band -fft -cheb  & 1039 & 2555 & 28.91\\ \hline
1005 & 2nd & -piste1 -silence -noise -band -aggr -diff  & 1025 & 2569 & 28.52\\ \hline
1006 & 2nd & -piste1 -silence -noise -band -aggr -cheb  & 1030 & 2564 & 28.66\\
\hline
\end{tabular}
\end{minipage}
\end{table}

\begin{table}[H]
\caption{Consolidated results (piste1-result1-marf-stats), Part 22.}
\label{tab:piste1-result1-marf-stats-results22}
\small
\begin{minipage}[b]{\textwidth}
\centering
\begin{tabular}{|c|c|l|c|c|r|} \hline
Run \# & Guess & Configuration & GOOD & BAD & Precision,\%\\ \hline\hline
1050 & 2nd & -piste1 -noise -norm -fft -mink  & 880 & 2714 & 24.49\\ \hline
1051 & 2nd & -piste1 -noise -norm -aggr -mink  & 880 & 2714 & 24.49\\ \hline
1052 & 2nd & -piste1 -low -fft -mink  & 880 & 2714 & 24.49\\ \hline
1053 & 2nd & -piste1 -low -aggr -mink  & 880 & 2714 & 24.49\\ \hline
1054 & 2nd & -piste1 -silence -noise -norm -aggr -mink  & 889 & 2705 & 24.74\\ \hline
1055 & 2nd & -piste1 -silence -low -fft -mink  & 889 & 2705 & 24.74\\ \hline
1056 & 2nd & -piste1 -silence -low -aggr -mink  & 889 & 2705 & 24.74\\
\hline
\end{tabular}
\end{minipage}
\end{table}

\begin{table}[H]
\caption{Consolidated results (piste1-result1-marf-stats), Part 23.}
\label{tab:piste1-result1-marf-stats-results23}
\small
\begin{minipage}[b]{\textwidth}
\centering
\begin{tabular}{|c|c|l|c|c|r|} \hline
Run \# & Guess & Configuration & GOOD & BAD & Precision,\%\\ \hline\hline
1100 & 2nd & -piste1 -silence -norm -lpc -eucl  & 679 & 2915 & 18.89\\ \hline
1101 & 2nd & -piste1 -norm -lpc -eucl  & 679 & 2915 & 18.89\\ \hline
1102 & 2nd & -piste1 -silence -noise -raw -lpc -eucl  & 679 & 2915 & 18.89\\ \hline
1103 & 2nd & -piste1 -noise -raw -lpc -eucl  & 679 & 2915 & 18.89\\ \hline
1104 & 2nd & -piste1 -raw -lpc -eucl  & 679 & 2915 & 18.89\\ \hline
1105 & 2nd & -piste1 -noise -high -fft -cheb  & 712 & 2882 & 19.81\\ \hline
1106 & 2nd & -piste1 -silence -raw -lpc -cos  & 671 & 2923 & 18.67\\
\hline
\end{tabular}
\end{minipage}
\end{table}

\begin{table}[H]
\caption{Consolidated results (piste1-result1-marf-stats), Part 24.}
\label{tab:piste1-result1-marf-stats-results24}
\small
\begin{minipage}[b]{\textwidth}
\centering
\begin{tabular}{|c|c|l|c|c|r|} \hline
Run \# & Guess & Configuration & GOOD & BAD & Precision,\%\\ \hline\hline
1150 & 2nd & -piste1 -silence -noise -endp -minmax -mink  & 659 & 2935 & 18.34\\ \hline
1151 & 2nd & -piste1 -noise -endp -lpc -cos  & 654 & 2940 & 18.20\\ \hline
1152 & 2nd & -piste1 -silence -noise -endp -lpc -diff  & 642 & 2952 & 17.86\\ \hline
1153 & 2nd & -piste1 -endp -minmax -mink  & 624 & 2970 & 17.36\\ \hline
1154 & 2nd & -piste1 -silence -endp -minmax -mink  & 624 & 2970 & 17.36\\ \hline
1155 & 2nd & -piste1 -noise -low -minmax -cheb  & 614 & 2980 & 17.08\\ \hline
1156 & 2nd & -piste1 -noise -endp -minmax -diff  & 607 & 2987 & 16.89\\
\hline
\end{tabular}
\end{minipage}
\end{table}

\begin{table}[H]
\caption{Consolidated results (piste1-result1-marf-stats), Part 25.}
\label{tab:piste1-result1-marf-stats-results25}
\small
\begin{minipage}[b]{\textwidth}
\centering
\begin{tabular}{|c|c|l|c|c|r|} \hline
Run \# & Guess & Configuration & GOOD & BAD & Precision,\%\\ \hline\hline
1200 & 2nd & -piste1 -noise -raw -minmax -cos  & 620 & 2974 & 17.25\\ \hline
1201 & 2nd & -piste1 -silence -endp -minmax -cos  & 615 & 2979 & 17.11\\ \hline
1202 & 2nd & -piste1 -silence -raw -minmax -cos  & 620 & 2974 & 17.25\\ \hline
1203 & 2nd & -piste1 -silence -noise -norm -minmax -eucl  & 647 & 2947 & 18.00\\ \hline
1204 & 2nd & -piste1 -silence -noise -raw -minmax -cos  & 620 & 2974 & 17.25\\ \hline
1205 & 2nd & -piste1 -raw -minmax -cos  & 620 & 2974 & 17.25\\ \hline
1206 & 2nd & -piste1 -norm -minmax -cos  & 623 & 2971 & 17.33\\
\hline
\end{tabular}
\end{minipage}
\end{table}

\begin{table}[H]
\caption{Consolidated results (piste1-result1-marf-stats), Part 26.}
\label{tab:piste1-result1-marf-stats-results26}
\small
\begin{minipage}[b]{\textwidth}
\centering
\begin{tabular}{|c|c|l|c|c|r|} \hline
Run \# & Guess & Configuration & GOOD & BAD & Precision,\%\\ \hline\hline
1250 & 2nd & -piste1 -bandstop -minmax -eucl  & 598 & 2996 & 16.64\\ \hline
1251 & 2nd & -piste1 -noise -high -aggr -eucl  & 604 & 2990 & 16.81\\ \hline
1252 & 2nd & -piste1 -noise -low -minmax -diff  & 590 & 3004 & 16.42\\ \hline
1253 & 2nd & -piste1 -silence -noise -high -fft -hamming  & 571 & 3023 & 15.89\\ \hline
1254 & 2nd & -piste1 -bandstop -minmax -diff  & 583 & 3011 & 16.22\\ \hline
1255 & 2nd & -piste1 -silence -noise -endp -lpc -mink  & 621 & 2973 & 17.28\\ \hline
1256 & 2nd & -piste1 -silence -endp -lpc -diff  & 617 & 2977 & 17.17\\
\hline
\end{tabular}
\end{minipage}
\end{table}

\begin{table}[H]
\caption{Consolidated results (piste1-result1-marf-stats), Part 27.}
\label{tab:piste1-result1-marf-stats-results27}
\small
\begin{minipage}[b]{\textwidth}
\centering
\begin{tabular}{|c|c|l|c|c|r|} \hline
Run \# & Guess & Configuration & GOOD & BAD & Precision,\%\\ \hline\hline
1300 & 2nd & -piste1 -noise -bandstop -minmax -diff  & 593 & 3001 & 16.50\\ \hline
1301 & 2nd & -piste1 -noise -low -lpc -cheb  & 543 & 3051 & 15.11\\ \hline
1302 & 2nd & -piste1 -noise -bandstop -minmax -eucl  & 599 & 2995 & 16.67\\ \hline
1303 & 2nd & -piste1 -noise -bandstop -minmax -mink  & 570 & 3024 & 15.86\\ \hline
1304 & 2nd & -piste1 -bandstop -lpc -cos  & 546 & 3048 & 15.19\\ \hline
1305 & 2nd & -piste1 -band -aggr -hamming  & 553 & 3041 & 15.39\\ \hline
1306 & 2nd & -piste1 -low -lpc -cheb  & 534 & 3060 & 14.86\\
\hline
\end{tabular}
\end{minipage}
\end{table}

\begin{table}[H]
\caption{Consolidated results (piste1-result1-marf-stats), Part 28.}
\label{tab:piste1-result1-marf-stats-results28}
\small
\begin{minipage}[b]{\textwidth}
\centering
\begin{tabular}{|c|c|l|c|c|r|} \hline
Run \# & Guess & Configuration & GOOD & BAD & Precision,\%\\ \hline\hline
1350 & 2nd & -piste1 -band -fft -hamming  & 549 & 3045 & 15.28\\ \hline
1351 & 2nd & -piste1 -silence -noise -low -lpc -cos  & 528 & 3066 & 14.69\\ \hline
1352 & 2nd & -piste1 -bandstop -lpc -mink  & 527 & 3067 & 14.66\\ \hline
1353 & 2nd & -piste1 -silence -noise -norm -minmax -hamming  & 517 & 3077 & 14.39\\ \hline
1354 & 2nd & -piste1 -silence -low -minmax -hamming  & 517 & 3077 & 14.39\\ \hline
1355 & 2nd & -piste1 -noise -bandstop -lpc -diff  & 526 & 3068 & 14.64\\ \hline
1356 & 2nd & -piste1 -band -lpc -diff  & 530 & 3064 & 14.75\\
\hline
\end{tabular}
\end{minipage}
\end{table}

\begin{table}[H]
\caption{Consolidated results (piste1-result1-marf-stats), Part 29.}
\label{tab:piste1-result1-marf-stats-results29}
\small
\begin{minipage}[b]{\textwidth}
\centering
\begin{tabular}{|c|c|l|c|c|r|} \hline
Run \# & Guess & Configuration & GOOD & BAD & Precision,\%\\ \hline\hline
1400 & 2nd & -piste1 -noise -low -fft -hamming  & 526 & 3068 & 14.64\\ \hline
1401 & 2nd & -piste1 -silence -low -lpc -mink  & 519 & 3075 & 14.44\\ \hline
1402 & 2nd & -piste1 -noise -band -lpc -diff  & 508 & 3086 & 14.13\\ \hline
1403 & 2nd & -piste1 -silence -noise -norm -lpc -mink  & 519 & 3075 & 14.44\\ \hline
1404 & 2nd & -piste1 -noise -band -lpc -eucl  & 525 & 3069 & 14.61\\ \hline
1405 & 2nd & -piste1 -silence -noise -high -minmax -cos  & 553 & 3041 & 15.39\\ \hline
1406 & 2nd & -piste1 -high -lpc -cos  & 519 & 3075 & 14.44\\
\hline
\end{tabular}
\end{minipage}
\end{table}

\begin{table}[H]
\caption{Consolidated results (piste1-result1-marf-stats), Part 30.}
\label{tab:piste1-result1-marf-stats-results30}
\small
\begin{minipage}[b]{\textwidth}
\centering
\begin{tabular}{|c|c|l|c|c|r|} \hline
Run \# & Guess & Configuration & GOOD & BAD & Precision,\%\\ \hline\hline
1450 & 2nd & -piste1 -silence -bandstop -lpc -hamming  & 496 & 3098 & 13.80\\ \hline
1451 & 2nd & -piste1 -noise -highpassboost -aggr -cheb  & 504 & 3090 & 14.02\\ \hline
1452 & 2nd & -piste1 -highpassboost -aggr -hamming  & 504 & 3090 & 14.02\\ \hline
1453 & 2nd & -piste1 -silence -noise -highpassboost -aggr -eucl  & 504 & 3090 & 14.02\\ \hline
1454 & 2nd & -piste1 -highpassboost -fft -mink  & 504 & 3090 & 14.02\\ \hline
1455 & 2nd & -piste1 -silence -noise -boost -aggr -diff  & 504 & 3090 & 14.02\\ \hline
1456 & 2nd & -piste1 -silence -noise -boost -aggr -cos  & 504 & 3090 & 14.02\\
\hline
\end{tabular}
\end{minipage}
\end{table}

\begin{table}[H]
\caption{Consolidated results (piste1-result1-marf-stats), Part 31.}
\label{tab:piste1-result1-marf-stats-results31}
\small
\begin{minipage}[b]{\textwidth}
\centering
\begin{tabular}{|c|c|l|c|c|r|} \hline
Run \# & Guess & Configuration & GOOD & BAD & Precision,\%\\ \hline\hline
1500 & 2nd & -piste1 -silence -noise -boost -lpc -diff  & 504 & 3090 & 14.02\\ \hline
1501 & 2nd & -piste1 -noise -highpassboost -aggr -eucl  & 504 & 3090 & 14.02\\ \hline
1502 & 2nd & -piste1 -boost -aggr -cheb  & 504 & 3090 & 14.02\\ \hline
1503 & 2nd & -piste1 -boost -aggr -cos  & 504 & 3090 & 14.02\\ \hline
1504 & 2nd & -piste1 -noise -boost -fft -diff  & 504 & 3090 & 14.02\\ \hline
1505 & 2nd & -piste1 -boost -minmax -cos  & 504 & 3090 & 14.02\\ \hline
1506 & 2nd & -piste1 -noise -boost -lpc -cheb  & 504 & 3090 & 14.02\\
\hline
\end{tabular}
\end{minipage}
\end{table}

\begin{table}[H]
\caption{Consolidated results (piste1-result1-marf-stats), Part 32.}
\label{tab:piste1-result1-marf-stats-results32}
\small
\begin{minipage}[b]{\textwidth}
\centering
\begin{tabular}{|c|c|l|c|c|r|} \hline
Run \# & Guess & Configuration & GOOD & BAD & Precision,\%\\ \hline\hline
1550 & 2nd & -piste1 -noise -boost -lpc -mink  & 504 & 3090 & 14.02\\ \hline
1551 & 2nd & -piste1 -silence -boost -aggr -hamming  & 504 & 3090 & 14.02\\ \hline
1552 & 2nd & -piste1 -silence -boost -lpc -eucl  & 504 & 3090 & 14.02\\ \hline
1553 & 2nd & -piste1 -silence -noise -highpassboost -aggr -diff  & 504 & 3090 & 14.02\\ \hline
1554 & 2nd & -piste1 -boost -lpc -cheb  & 504 & 3090 & 14.02\\ \hline
1555 & 2nd & -piste1 -silence -noise -boost -lpc -eucl  & 504 & 3090 & 14.02\\ \hline
1556 & 2nd & -piste1 -noise -boost -minmax -hamming  & 504 & 3090 & 14.02\\
\hline
\end{tabular}
\end{minipage}
\end{table}

\begin{table}[H]
\caption{Consolidated results (piste1-result1-marf-stats), Part 33.}
\label{tab:piste1-result1-marf-stats-results33}
\small
\begin{minipage}[b]{\textwidth}
\centering
\begin{tabular}{|c|c|l|c|c|r|} \hline
Run \# & Guess & Configuration & GOOD & BAD & Precision,\%\\ \hline\hline
1600 & 2nd & -piste1 -noise -highpassboost -aggr -diff  & 504 & 3090 & 14.02\\ \hline
1601 & 2nd & -piste1 -boost -lpc -eucl  & 504 & 3090 & 14.02\\ \hline
1602 & 2nd & -piste1 -highpassboost -lpc -mink  & 504 & 3090 & 14.02\\ \hline
1603 & 2nd & -piste1 -silence -highpassboost -aggr -cheb  & 504 & 3090 & 14.02\\ \hline
1604 & 2nd & -piste1 -silence -highpassboost -fft -cheb  & 504 & 3090 & 14.02\\ \hline
1605 & 2nd & -piste1 -silence -highpassboost -aggr -hamming  & 504 & 3090 & 14.02\\ \hline
1606 & 2nd & -piste1 -silence -noise -highpassboost -fft -cos  & 504 & 3090 & 14.02\\
\hline
\end{tabular}
\end{minipage}
\end{table}

\begin{table}[H]
\caption{Consolidated results (piste1-result1-marf-stats), Part 34.}
\label{tab:piste1-result1-marf-stats-results34}
\small
\begin{minipage}[b]{\textwidth}
\centering
\begin{tabular}{|c|c|l|c|c|r|} \hline
Run \# & Guess & Configuration & GOOD & BAD & Precision,\%\\ \hline\hline
1650 & 2nd & -piste1 -silence -norm -aggr -hamming  & 459 & 3135 & 12.77\\ \hline
1651 & 2nd & -piste1 -silence -endp -aggr -hamming  & 480 & 3114 & 13.36\\ \hline
1652 & 2nd & -piste1 -endp -aggr -hamming  & 480 & 3114 & 13.36\\ \hline
1653 & 2nd & -piste1 -silence -noise -endp -aggr -hamming  & 486 & 3108 & 13.52\\ \hline
1654 & 2nd & -piste1 -silence -noise -band -lpc -hamming  & 506 & 3088 & 14.08\\ \hline
1655 & 2nd & -piste1 -silence -noise -raw -aggr -hamming  & 460 & 3134 & 12.80\\ \hline
1656 & 2nd & -piste1 -noise -raw -aggr -hamming  & 460 & 3134 & 12.80\\ \hline
1657 & 2nd & -piste1 -raw -aggr -hamming  & 460 & 3134 & 12.80\\ \hline
1658 & 2nd & -piste1 -silence -raw -aggr -hamming  & 460 & 3134 & 12.80\\ \hline
1659 & 2nd & -piste1 -noise -band -lpc -hamming  & 457 & 3137 & 12.72\\ \hline
1660 & 2nd & -piste1 -silence -low -fft -hamming  & 438 & 3156 & 12.19\\ \hline
1661 & 2nd & -piste1 -silence -noise -norm -fft -hamming  & 438 & 3156 & 12.19\\ \hline
1662 & 2nd & -piste1 -raw -fft -hamming  & 454 & 3140 & 12.63\\ \hline
1663 & 2nd & -piste1 -silence -raw -fft -hamming  & 454 & 3140 & 12.63\\ \hline
1664 & 2nd & -piste1 -noise -raw -fft -hamming  & 454 & 3140 & 12.63\\ \hline
1665 & 2nd & -piste1 -silence -noise -high -lpc -hamming  & 469 & 3125 & 13.05\\ \hline
1666 & 2nd & -piste1 -silence -noise -raw -fft -hamming  & 454 & 3140 & 12.63\\ \hline
1667 & 2nd & -piste1 -silence -noise -norm -aggr -hamming  & 466 & 3128 & 12.97\\ \hline
1668 & 2nd & -piste1 -silence -low -aggr -hamming  & 466 & 3128 & 12.97\\ \hline
\end{tabular}
\end{minipage}
\end{table}

\begin{table}[H]
\caption{Consolidated results (piste1-result1-marf-stats), Part 35.}
\label{tab:piste1-result1-marf-stats-results35}
\small
\begin{minipage}[b]{\textwidth}
\centering
\begin{tabular}{|c|c|l|c|c|r|} \hline
Run \# & Guess & Configuration & GOOD & BAD & Precision,\%\\ \hline\hline
1 & 1st & 1830 & 84722 & 125446 & 40.31\\ \hline
2 & 1st & 1940 & 38330 & 147652 & 20.61\\ \hline
3 & 1st & 1810 & 38811 & 171357 & 18.47\\ \hline
4 & 1st & 1820 & 29107 & 181061 & 13.85\\ \hline
5 & 1st & 1900 & 15229 & 166583 & 8.38\\ \hline
6 & 1st & 1920 & 15360 & 168954 & 8.33\\ \hline
7 & 1st & 1850 & 16807 & 193361 & 8.00\\ \hline
8 & 1st & 1880 & 15643 & 194525 & 7.44\\ \hline
9 & 1st & 1800 & 14974 & 195194 & 7.12\\ \hline
10 & 1st & 1860 & 14038 & 195296 & 6.71\\ \hline
11 & 1st & 1870 & 14037 & 196131 & 6.68\\ \hline
12 & 1st & 1930 & 11314 & 173000 & 6.14\\ \hline
13 & 1st & 1840 & 12468 & 197700 & 5.93\\ \hline
14 & 1st & 1890 & 10727 & 176089 & 5.74\\ \hline
15 & 1st & 1910 & 10408 & 173072 & 5.67\\ \hline
16 & 2nd & 1830 & 105137 & 105031 & 50.03\\ \hline
17 & 2nd & 1940 & 53906 & 132076 & 28.98\\ \hline
18 & 2nd & 1810 & 98904 & 111264 & 47.06\\ \hline
19 & 2nd & 1820 & 45219 & 164949 & 21.52\\ \hline
20 & 2nd & 1900 & 28690 & 153122 & 15.78\\ \hline
21 & 2nd & 1920 & 30199 & 154115 & 16.38\\ \hline
22 & 2nd & 1850 & 32345 & 177823 & 15.39\\ \hline
23 & 2nd & 1880 & 36923 & 173245 & 17.57\\ \hline
24 & 2nd & 1800 & 38997 & 171171 & 18.56\\ \hline
25 & 2nd & 1860 & 29063 & 180271 & 13.88\\ \hline
26 & 2nd & 1870 & 28356 & 181812 & 13.49\\ \hline
27 & 2nd & 1930 & 23456 & 160858 & 12.73\\ \hline
28 & 2nd & 1840 & 25489 & 184679 & 12.13\\ \hline
29 & 2nd & 1890 & 21281 & 165535 & 11.39\\ \hline
30 & 2nd & 1910 & 19657 & 163823 & 10.71\\ \hline
\end{tabular}
\end{minipage}
\end{table}

\begin{table}[H]
\caption{Consolidated results (piste1-result3-nlp), Part 1.}
\label{tab:piste1-result3-nlp-results1}
\small
\begin{minipage}[b]{\textwidth}
\centering
\begin{tabular}{|c|c|l|c|c|r|} \hline
Run \# & Guess & Configuration & GOOD & BAD & Precision,\%\\ \hline\hline
1 & 1st & -piste1 -nopreprep -char -unigram -add-delta  & 400 & 3194 & 11.13\\ \hline
2 & 1st & -piste1 -nopreprep -char -unigram -mle  & 386 & 3208 & 10.74\\ \hline
3 & 1st & -piste1 -nopreprep -char -unigram -witten-bell  & 354 & 3240 & 9.85\\ \hline
4 & 1st & -piste1 -nopreprep -char -unigram -add-one  & 344 & 3250 & 9.57\\ \hline
5 & 1st & -piste1 -nopreprep -char -unigram -good-turing  & 315 & 3279 & 8.76\\ \hline
6 & 1st & -piste1 -nopreprep -char -bigram -add-one  & 266 & 3328 & 7.40\\ \hline
7 & 1st & -piste1 -nopreprep -char -bigram -add-delta  & 265 & 3329 & 7.37\\ \hline
8 & 1st & -piste1 -nopreprep -char -bigram -good-turing  & 254 & 3340 & 7.07\\ \hline
9 & 1st & -piste1 -nopreprep -char -bigram -witten-bell  & 252 & 3342 & 7.01\\ \hline
10 & 1st & -piste1 -nopreprep -char -bigram -mle  & 252 & 3342 & 7.01\\ \hline
11 & 2nd & -piste1 -nopreprep -char -unigram -add-delta  & 692 & 2902 & 19.25\\ \hline
12 & 2nd & -piste1 -nopreprep -char -unigram -mle  & 756 & 2838 & 21.04\\ \hline
13 & 2nd & -piste1 -nopreprep -char -unigram -witten-bell  & 611 & 2983 & 17.00\\ \hline
14 & 2nd & -piste1 -nopreprep -char -unigram -add-one  & 656 & 2938 & 18.25\\ \hline
15 & 2nd & -piste1 -nopreprep -char -unigram -good-turing  & 547 & 3047 & 15.22\\ \hline
16 & 2nd & -piste1 -nopreprep -char -bigram -add-one  & 542 & 3052 & 15.08\\ \hline
17 & 2nd & -piste1 -nopreprep -char -bigram -add-delta  & 551 & 3043 & 15.33\\ \hline
18 & 2nd & -piste1 -nopreprep -char -bigram -good-turing  & 553 & 3041 & 15.39\\ \hline
19 & 2nd & -piste1 -nopreprep -char -bigram -witten-bell  & 504 & 3090 & 14.02\\ \hline
20 & 2nd & -piste1 -nopreprep -char -bigram -mle  & 504 & 3090 & 14.02\\ \hline
\end{tabular}
\end{minipage}
\end{table}

\begin{table}[H]
\caption{Consolidated results (piste1-result3-nlp), Part 2.}
\label{tab:piste1-result3-nlp-results2}
\small
\begin{minipage}[b]{\textwidth}
\centering
\begin{tabular}{|c|c|l|c|c|r|} \hline
Run \# & Guess & Configuration & GOOD & BAD & Precision,\%\\ \hline\hline
1 & 1st & 1830 & 707 & 1813 & 28.06\\ \hline
2 & 1st & 1920 & 453 & 1757 & 20.50\\ \hline
3 & 1st & 1820 & 445 & 2075 & 17.66\\ \hline
4 & 1st & 1930 & 339 & 1871 & 15.34\\ \hline
5 & 1st & 1890 & 246 & 1994 & 10.98\\ \hline
6 & 1st & 1800 & 230 & 2290 & 9.13\\ \hline
7 & 1st & 1810 & 151 & 2369 & 5.99\\ \hline
8 & 1st & 1870 & 140 & 2380 & 5.56\\ \hline
9 & 1st & 1940 & 93 & 2137 & 4.17\\ \hline
10 & 1st & 1860 & 94 & 2416 & 3.75\\ \hline
11 & 1st & 1910 & 52 & 2148 & 2.36\\ \hline
12 & 1st & 1900 & 43 & 2137 & 1.97\\ \hline
13 & 1st & 1880 & 41 & 2479 & 1.63\\ \hline
14 & 1st & 1850 & 32 & 2488 & 1.27\\ \hline
15 & 1st & 1840 & 22 & 2498 & 0.87\\ \hline
16 & 2nd & 1830 & 924 & 1596 & 36.67\\ \hline
17 & 2nd & 1920 & 466 & 1744 & 21.09\\ \hline
18 & 2nd & 1820 & 605 & 1915 & 24.01\\ \hline
19 & 2nd & 1930 & 447 & 1763 & 20.23\\ \hline
20 & 2nd & 1890 & 286 & 1954 & 12.77\\ \hline
21 & 2nd & 1800 & 493 & 2027 & 19.56\\ \hline
22 & 2nd & 1810 & 1056 & 1464 & 41.90\\ \hline
23 & 2nd & 1870 & 306 & 2214 & 12.14\\ \hline
24 & 2nd & 1940 & 540 & 1690 & 24.22\\ \hline
25 & 2nd & 1860 & 243 & 2267 & 9.68\\ \hline
26 & 2nd & 1910 & 144 & 2056 & 6.55\\ \hline
27 & 2nd & 1900 & 140 & 2040 & 6.42\\ \hline
28 & 2nd & 1880 & 103 & 2417 & 4.09\\ \hline
29 & 2nd & 1850 & 102 & 2418 & 4.05\\ \hline
30 & 2nd & 1840 & 61 & 2459 & 2.42\\ \hline
\end{tabular}
\end{minipage}
\end{table}

\subsection{Piste 2 / Track 2: Place of Origin}
\label{sect:results-piste2}
\index{Results!Piste 2}
\index{Results!Track 2}

France vs. Quebec; Journals

\subsubsection{Results for testing data for text-only processing}
\index{Results!testing data for text-only processing}

\begin{table}[H]
\caption{Consolidated results (piste2-result1-text-only-eval), Part 1.}
\label{tab:piste2-result1-text-only-eval-results1}
\small
\begin{minipage}[b]{\textwidth}
\centering
\begin{tabular}{|c|c|l|c|c|r|} \hline
Run \# & Guess & Configuration & GOOD & BAD & Precision,\%\\ \hline\hline
1 & 1st & -text-only -ref -noise -band -fft -diff  & 1392 & 1090 & 56.08\\ \hline
2 & 1st & -text-only -ref -noise -band -aggr -diff  & 1391 & 1091 & 56.04\\ \hline
3 & 1st & -text-only -ref -noise -band -fft -cheb  & 1386 & 1096 & 55.84\\ \hline
4 & 1st & -text-only -ref -noise -band -aggr -cheb  & 1386 & 1096 & 55.84\\ \hline
5 & 1st & -text-only -ref -high -fft -diff  & 1380 & 1102 & 55.60\\ \hline
6 & 1st & -text-only -ref -band -aggr -cheb  & 1380 & 1102 & 55.60\\ \hline
7 & 1st & -text-only -ref -band -fft -cheb  & 1379 & 1103 & 55.56\\
\hline
\end{tabular}
\end{minipage}
\end{table}

\begin{table}[H]
\caption{Consolidated results (piste2-result1-text-only-eval), Part 2.}
\label{tab:piste2-result1-text-only-eval-results2}
\small
\begin{minipage}[b]{\textwidth}
\centering
\begin{tabular}{|c|c|l|c|c|r|} \hline
Run \# & Guess & Configuration & GOOD & BAD & Precision,\%\\ \hline\hline
50 & 2nd & -text-only -ref -noise -band -fft -diff  & 2482 & 0 & 100.00\\ \hline
51 & 2nd & -text-only -ref -noise -band -aggr -diff  & 2482 & 0 & 100.00\\ \hline
52 & 2nd & -text-only -ref -noise -band -fft -cheb  & 2482 & 0 & 100.00\\ \hline
53 & 2nd & -text-only -ref -noise -band -aggr -cheb  & 2482 & 0 & 100.00\\ \hline
54 & 2nd & -text-only -ref -high -fft -diff  & 2482 & 0 & 100.00\\ \hline
55 & 2nd & -text-only -ref -band -aggr -cheb  & 2482 & 0 & 100.00\\ \hline
56 & 2nd & -text-only -ref -band -fft -cheb  & 2482 & 0 & 100.00\\
\hline
\end{tabular}
\end{minipage}
\end{table}

\begin{table}[H]
\caption{Consolidated results (piste2-result1-text-only-eval), Part 3.}
\label{tab:piste2-result1-text-only-eval-results3}
\small
\begin{minipage}[b]{\textwidth}
\centering
\begin{tabular}{|c|c|l|c|c|r|} \hline
Run \# & Guess & Configuration & GOOD & BAD & Precision,\%\\ \hline\hline
1 & 1st & France & 32445 & 24052 & 57.43\\ \hline
2 & 1st & Quebec & 31720 & 33401 & 48.71\\ \hline
3 & 2nd & France & 56497 & 0 & 100.00\\ \hline
4 & 2nd & Quebec & 65121 & 0 & 100.00\\ \hline
\end{tabular}
\end{minipage}
\end{table}

\begin{table}[H]
\caption{Fichier \'{e}valu\'{e} : \tiny{\url{equipe_3_tache_2_execution_3-testing.sh-text-only-ref-band-aggr-cheb.prexml.xml}}}
\label{tab:equipe_3_tache_2_execution_3-testing.sh-text-only-ref-band-aggr-cheb.prexml.xml}
\small
\begin{minipage}[b]{\textwidth}
\centering
\begin{tabular}{|l|} \hline
\hline Evaluation du pays\\\hline\hline
\verb+- classe F (attendus = 1153, ramenes = 1491.00, corrects = 771.00)+\\
\verb+  rappel = 0.669	precision = 0.517	f-mesure = 0.583+\\\hline
\verb+- classe Q (attendus = 1329, ramenes = 991.00, corrects = 609.00)+\\
\verb+  rappel = 0.458	precision = 0.615	f-mesure = 0.525+\\\hline
\verb+- sur l'ensemble des 2 classes+\\
\verb+  macro rappel = 0.563	macro precision = 0.566	macro F-mesure = 0.565+\\\hline
\hline Evaluation du journal\\\hline\hline
\verb+- classe D (attendus = 652, ramenes = 0.00, corrects = 0.00)+\\
\verb+  rappel = 0.000	precision = 0.000	f-mesure = 0.000+\\\hline
\verb+- classe E (attendus = 553, ramenes = 0.00, corrects = 0.00)+\\
\verb+  rappel = 0.000	precision = 0.000	f-mesure = 0.000+\\\hline
\verb+- classe M (attendus = 600, ramenes = 1491.00, corrects = 370.00)+\\
\verb+  rappel = 0.617	precision = 0.248	f-mesure = 0.354+\\\hline
\verb+- classe P (attendus = 677, ramenes = 991.00, corrects = 322.00)+\\
\verb+  rappel = 0.476	precision = 0.325	f-mesure = 0.386+\\\hline
\verb+- sur l'ensemble des 4 classes+\\
\verb+  macro rappel = 0.273	macro precision = 0.143	macro F-mesure = 0.188+\\\hline
\hline
\end{tabular}
\end{minipage}
\end{table}
\begin{table}[H]
\caption{Fichier \'{e}valu\'{e} : \tiny{\url{equipe_3_tache_2_execution_3-testing.sh-text-only-ref-band-aggr-cos.prexml.xml}}}
\label{tab:equipe_3_tache_2_execution_3-testing.sh-text-only-ref-band-aggr-cos.prexml.xml}
\small
\begin{minipage}[b]{\textwidth}
\centering
\begin{tabular}{|l|} \hline
\hline Evaluation du pays\\\hline\hline
\verb+- classe F (attendus = 1153, ramenes = 1235.00, corrects = 598.00)+\\
\verb+  rappel = 0.519	precision = 0.484	f-mesure = 0.501+\\\hline
\verb+- classe Q (attendus = 1329, ramenes = 1247.00, corrects = 692.00)+\\
\verb+  rappel = 0.521	precision = 0.555	f-mesure = 0.537+\\\hline
\verb+- sur l'ensemble des 2 classes+\\
\verb+  macro rappel = 0.520	macro precision = 0.520	macro F-mesure = 0.520+\\\hline
\hline Evaluation du journal\\\hline\hline
\verb+- classe D (attendus = 652, ramenes = 0.00, corrects = 0.00)+\\
\verb+  rappel = 0.000	precision = 0.000	f-mesure = 0.000+\\\hline
\verb+- classe E (attendus = 553, ramenes = 0.00, corrects = 0.00)+\\
\verb+  rappel = 0.000	precision = 0.000	f-mesure = 0.000+\\\hline
\verb+- classe M (attendus = 600, ramenes = 1235.00, corrects = 312.00)+\\
\verb+  rappel = 0.520	precision = 0.253	f-mesure = 0.340+\\\hline
\verb+- classe P (attendus = 677, ramenes = 1247.00, corrects = 358.00)+\\
\verb+  rappel = 0.529	precision = 0.287	f-mesure = 0.372+\\\hline
\verb+- sur l'ensemble des 4 classes+\\
\verb+  macro rappel = 0.262	macro precision = 0.135	macro F-mesure = 0.178+\\\hline
\hline
\end{tabular}
\end{minipage}
\end{table}
\begin{table}[H]
\caption{Fichier \'{e}valu\'{e} : \tiny{\url{equipe_3_tache_2_execution_3-testing.sh-text-only-ref-band-aggr-diff.prexml.xml}}}
\label{tab:equipe_3_tache_2_execution_3-testing.sh-text-only-ref-band-aggr-diff.prexml.xml}
\small
\begin{minipage}[b]{\textwidth}
\centering
%
\end{minipage}
\end{table}
\begin{table}[H]
\caption{Fichier \'{e}valu\'{e} : \tiny{\url{equipe_3_tache_2_execution_3-testing.sh-text-only-ref-band-aggr-eucl.prexml.xml}}}
\label{tab:equipe_3_tache_2_execution_3-testing.sh-text-only-ref-band-aggr-eucl.prexml.xml}
\small
\begin{minipage}[b]{\textwidth}
\centering
%
\end{minipage}
\end{table}
\begin{table}[H]
\caption{Fichier \'{e}valu\'{e} : \tiny{\url{equipe_3_tache_2_execution_3-testing.sh-text-only-ref-band-fft-cheb.prexml.xml}}}
\label{tab:equipe_3_tache_2_execution_3-testing.sh-text-only-ref-band-fft-cheb.prexml.xml}
\small
\begin{minipage}[b]{\textwidth}
\centering
%
\end{minipage}
\end{table}
\begin{table}[H]
\caption{Fichier \'{e}valu\'{e} : \tiny{\url{equipe_3_tache_2_execution_3-testing.sh-text-only-ref-band-fft-cos.prexml.xml}}}
\label{tab:equipe_3_tache_2_execution_3-testing.sh-text-only-ref-band-fft-cos.prexml.xml}
\small
\begin{minipage}[b]{\textwidth}
\centering
%
\end{minipage}
\end{table}
\begin{table}[H]
\caption{Fichier \'{e}valu\'{e} : \tiny{\url{equipe_3_tache_2_execution_3-testing.sh-text-only-ref-band-fft-diff.prexml.xml}}}
\label{tab:equipe_3_tache_2_execution_3-testing.sh-text-only-ref-band-fft-diff.prexml.xml}
\small
\begin{minipage}[b]{\textwidth}
\centering
%
\end{minipage}
\end{table}

\begin{table}[H]
\caption{Consolidated results (piste2-result1-marf-text-only-journal-eval), Part 1.}
\label{tab:piste2-result1-marf-text-only-journal-eval-results1}
\small
\begin{minipage}[b]{\textwidth}
\centering
%
\end{minipage}
\end{table}

\subsubsection{Results for training data for text-only processing}
\index{Results!training data for text-only processing}

\begin{table}[H]
\caption{Consolidated results (piste2-result1-marf-text-only-stats), Part 1.}
\label{tab:piste2-result1-marf-text-only-stats-results1}
\small
\begin{minipage}[b]{\textwidth}
\centering
%
\end{minipage}
\caption{Consolidated results (piste2-result1-marf-text-only-journal), Part 23.}
\label{tab:piste2-result1-marf-text-only-journal-results23}
\end{table}


\subsubsection{Results for testing data for title+text processing}
\index{Results!testing data for title+text processing}

\begin{table}[H]
\caption{Consolidated results (piste2-result1-marf-text-title-eval), Part 1.}
\label{tab:piste2-result1-marf-text-title-eval-results1}
\small
\begin{minipage}[b]{\textwidth}
\centering
%
\end{minipage}
\end{table}

\subsubsection{Results for training data for title+text processing}
\index{Results!training data for title+text processing}

\begin{table}[H]
\caption{Consolidated results (piste2-result1-marf-text-title-stats), Part 1.}
\label{tab:piste2-result1-marf-text-title-stats-results1}
\small
\begin{minipage}[b]{\textwidth}
\centering
%
\end{minipage}
\caption{Consolidated results (piste2-result1-marf-title-only-journal-slow2), Part 22.}
\label{tab:piste2-result1-marf-title-only-journal-slow2-results22}
\end{table}


\subsubsection{Results for testing data for title-only processing}
\index{Results!testing data for title-only processing}

\begin{table}[H]
\caption{Consolidated results (piste2-result1-marf-title-only-eval), Part 1.}
\label{tab:piste2-result1-marf-title-only-eval-results1}
\small
\begin{minipage}[b]{\textwidth}
\centering

\end{minipage}
\end{table}

\begin{table}[H]
\caption{Consolidated results (piste2-result1-marf-title-only-eval), Part 2.}
\label{tab:piste2-result1-marf-title-only-eval-results2}
\small
\begin{minipage}[b]{\textwidth}
\centering
%
\end{minipage}
\end{table}

\begin{table}[H]
\caption{Consolidated results (piste2-result1-marf-title-only-eval), Part 3.}
\label{tab:piste2-result1-marf-title-only-eval-results3}
\small
\begin{minipage}[b]{\textwidth}
\centering
%
\end{minipage}
\end{table}

\begin{table}[H]
\caption{Fichier \'{e}valu\'{e} : \tiny{\url{equipe_3_tache_2_execution_3-testing.sh-title-only-ref-bandstop-aggr-cos.prexml.xml}}}
\label{tab:equipe_3_tache_2_execution_3-testing.sh-title-only-ref-bandstop-aggr-cos.prexml.xml}
\small
\begin{minipage}[b]{\textwidth}
\centering
%
\end{minipage}
\end{table}
\begin{table}[H]
\caption{Fichier \'{e}valu\'{e} : \tiny{\url{equipe_3_tache_2_execution_3-testing.sh-title-only-ref-bandstop-aggr-eucl.prexml.xml}}}
\label{tab:equipe_3_tache_2_execution_3-testing.sh-title-only-ref-bandstop-aggr-eucl.prexml.xml}
\small
\begin{minipage}[b]{\textwidth}
\centering
%
\end{minipage}
\end{table}
\begin{table}[H]
\caption{Fichier \'{e}valu\'{e} : \tiny{\url{equipe_3_tache_2_execution_3-testing.sh-title-only-ref-bandstop-fft-cos.prexml.xml}}}
\label{tab:equipe_3_tache_2_execution_3-testing.sh-title-only-ref-bandstop-fft-cos.prexml.xml}
\small
\begin{minipage}[b]{\textwidth}
\centering
%
\end{minipage}
\end{table}
\begin{table}[H]
\caption{Fichier \'{e}valu\'{e} : \tiny{\url{equipe_3_tache_2_execution_3-testing.sh-title-only-ref-bandstop-fft-eucl.prexml.xml}}}
\label{tab:equipe_3_tache_2_execution_3-testing.sh-title-only-ref-bandstop-fft-eucl.prexml.xml}
\small
\begin{minipage}[b]{\textwidth}
\centering
%
\end{minipage}
\end{table}
\begin{table}[H]
\caption{Fichier \'{e}valu\'{e} : \tiny{\url{equipe_3_tache_2_execution_3-testing.sh-title-only-ref-endp-fft-cos.prexml.xml}}}
\label{tab:equipe_3_tache_2_execution_3-testing.sh-title-only-ref-endp-fft-cos.prexml.xml}
\small
\begin{minipage}[b]{\textwidth}
\centering
%
\end{minipage}
\end{table}
\begin{table}[H]
\caption{Fichier \'{e}valu\'{e} : \tiny{\url{equipe_3_tache_2_execution_3-testing.sh-title-only-ref-low-aggr-eucl.prexml.xml}}}
\label{tab:equipe_3_tache_2_execution_3-testing.sh-title-only-ref-low-aggr-eucl.prexml.xml}
\small
\begin{minipage}[b]{\textwidth}
\centering
%
\end{minipage}
\end{table}
\begin{table}[H]
\caption{Fichier \'{e}valu\'{e} : \tiny{\url{equipe_3_tache_2_execution_3-testing.sh-title-only-ref-low-fft-eucl.prexml.xml}}}
\label{tab:equipe_3_tache_2_execution_3-testing.sh-title-only-ref-low-fft-eucl.prexml.xml}
\small
\begin{minipage}[b]{\textwidth}
\centering
%
\end{minipage}
\end{table}
\begin{table}[H]
\caption{Fichier \'{e}valu\'{e} : \tiny{\url{equipe_3_tache_2_execution_3-testing.sh-title-only-ref-noise-bandstop-aggr-eucl.prexml.xml}}}
\label{tab:equipe_3_tache_2_execution_3-testing.sh-title-only-ref-noise-bandstop-aggr-eucl.prexml.xml}
\small
\begin{minipage}[b]{\textwidth}
\centering
%
\end{minipage}
\end{table}
\begin{table}[H]
\caption{Fichier \'{e}valu\'{e} : \tiny{\url{equipe_3_tache_2_execution_3-testing.sh-title-only-ref-noise-bandstop-fft-eucl.prexml.xml}}}
\label{tab:equipe_3_tache_2_execution_3-testing.sh-title-only-ref-noise-bandstop-fft-eucl.prexml.xml}
\small
\begin{minipage}[b]{\textwidth}
\centering
%
\end{minipage}
\end{table}
\begin{table}[H]
\caption{Fichier \'{e}valu\'{e} : \tiny{\url{equipe_3_tache_2_execution_3-testing.sh-title-only-ref-noise-endp-aggr-eucl.prexml.xml}}}
\label{tab:equipe_3_tache_2_execution_3-testing.sh-title-only-ref-noise-endp-aggr-eucl.prexml.xml}
\small
\begin{minipage}[b]{\textwidth}
\centering
%
\end{minipage}
\end{table}
\begin{table}[H]
\caption{Fichier \'{e}valu\'{e} : \tiny{\url{equipe_3_tache_2_execution_3-testing.sh-title-only-ref-noise-endp-fft-eucl.prexml.xml}}}
\label{tab:equipe_3_tache_2_execution_3-testing.sh-title-only-ref-noise-endp-fft-eucl.prexml.xml}
\small
\begin{minipage}[b]{\textwidth}
\centering
%
\end{minipage}
\end{table}
\begin{table}[H]
\caption{Fichier \'{e}valu\'{e} : \tiny{\url{equipe_3_tache_2_execution_3-testing.sh-title-only-ref-noise-low-aggr-eucl.prexml.xml}}}
\label{tab:equipe_3_tache_2_execution_3-testing.sh-title-only-ref-noise-low-aggr-eucl.prexml.xml}
\small
\begin{minipage}[b]{\textwidth}
\centering
%
\end{minipage}
\end{table}
\begin{table}[H]
\caption{Fichier \'{e}valu\'{e} : \tiny{\url{equipe_3_tache_2_execution_3-testing.sh-title-only-ref-noise-low-fft-eucl.prexml.xml}}}
\label{tab:equipe_3_tache_2_execution_3-testing.sh-title-only-ref-noise-low-fft-eucl.prexml.xml}
\small
\begin{minipage}[b]{\textwidth}
\centering
%
\end{minipage}
\end{table}
\begin{table}[H]
\caption{Fichier \'{e}valu\'{e} : \tiny{\url{equipe_3_tache_2_execution_3-testing.sh-title-only-ref-noise-norm-aggr-eucl.prexml.xml}}}
\label{tab:equipe_3_tache_2_execution_3-testing.sh-title-only-ref-noise-norm-aggr-eucl.prexml.xml}
\small
\begin{minipage}[b]{\textwidth}
\centering
%
\end{minipage}
\end{table}
\begin{table}[H]
\caption{Fichier \'{e}valu\'{e} : \tiny{\url{equipe_3_tache_2_execution_3-testing.sh-title-only-ref-noise-norm-fft-eucl.prexml.xml}}}
\label{tab:equipe_3_tache_2_execution_3-testing.sh-title-only-ref-noise-norm-fft-eucl.prexml.xml}
\small
\begin{minipage}[b]{\textwidth}
\centering
%
\end{minipage}
\end{table}
\begin{table}[H]
\caption{Fichier \'{e}valu\'{e} : \tiny{\url{equipe_3_tache_2_execution_3-testing.sh-title-only-ref-noise-raw-aggr-eucl.prexml.xml}}}
\label{tab:equipe_3_tache_2_execution_3-testing.sh-title-only-ref-noise-raw-aggr-eucl.prexml.xml}
\small
\begin{minipage}[b]{\textwidth}
\centering
%
\end{minipage}
\end{table}
\begin{table}[H]
\caption{Fichier \'{e}valu\'{e} : \tiny{\url{equipe_3_tache_2_execution_3-testing.sh-title-only-ref-noise-raw-fft-eucl.prexml.xml}}}
\label{tab:equipe_3_tache_2_execution_3-testing.sh-title-only-ref-noise-raw-fft-eucl.prexml.xml}
\small
\begin{minipage}[b]{\textwidth}
\centering
%
\end{minipage}
\end{table}
\begin{table}[H]
\caption{Fichier \'{e}valu\'{e} : \tiny{\url{equipe_3_tache_2_execution_3-testing.sh-title-only-ref-norm-aggr-cos.prexml.xml}}}
\label{tab:equipe_3_tache_2_execution_3-testing.sh-title-only-ref-norm-aggr-cos.prexml.xml}
\small
\begin{minipage}[b]{\textwidth}
\centering
%
\end{minipage}
\end{table}
\begin{table}[H]
\caption{Fichier \'{e}valu\'{e} : \tiny{\url{equipe_3_tache_2_execution_3-testing.sh-title-only-ref-norm-aggr-eucl.prexml.xml}}}
\label{tab:equipe_3_tache_2_execution_3-testing.sh-title-only-ref-norm-aggr-eucl.prexml.xml}
\small
\begin{minipage}[b]{\textwidth}
\centering
%
\end{minipage}
\end{table}
\begin{table}[H]
\caption{Fichier \'{e}valu\'{e} : \tiny{\url{equipe_3_tache_2_execution_3-testing.sh-title-only-ref-norm-fft-cos.prexml.xml}}}
\label{tab:equipe_3_tache_2_execution_3-testing.sh-title-only-ref-norm-fft-cos.prexml.xml}
\small
\begin{minipage}[b]{\textwidth}
\centering
%
\end{minipage}
\end{table}
\begin{table}[H]
\caption{Fichier \'{e}valu\'{e} : \tiny{\url{equipe_3_tache_2_execution_3-testing.sh-title-only-ref-norm-fft-eucl.prexml.xml}}}
\label{tab:equipe_3_tache_2_execution_3-testing.sh-title-only-ref-norm-fft-eucl.prexml.xml}
\small
\begin{minipage}[b]{\textwidth}
\centering
%
\end{minipage}
\end{table}
\begin{table}[H]
\caption{Fichier \'{e}valu\'{e} : \tiny{\url{equipe_3_tache_2_execution_3-testing.sh-title-only-ref-raw-aggr-eucl.prexml.xml}}}
\label{tab:equipe_3_tache_2_execution_3-testing.sh-title-only-ref-raw-aggr-eucl.prexml.xml}
\small
\begin{minipage}[b]{\textwidth}
\centering
%
\end{minipage}
\end{table}
\begin{table}[H]
\caption{Fichier \'{e}valu\'{e} : \tiny{\url{equipe_3_tache_2_execution_3-testing.sh-title-only-ref-raw-fft-eucl.prexml.xml}}}
\label{tab:equipe_3_tache_2_execution_3-testing.sh-title-only-ref-raw-fft-eucl.prexml.xml}
\small
\begin{minipage}[b]{\textwidth}
\centering
%
\end{minipage}
\end{table}
\begin{table}[H]
\caption{Fichier \'{e}valu\'{e} : \tiny{\url{equipe_3_tache_2_execution_3-testing.sh-title-only-ref-silence-bandstop-aggr-cos.prexml.xml}}}
\label{tab:equipe_3_tache_2_execution_3-testing.sh-title-only-ref-silence-bandstop-aggr-cos.prexml.xml}
\small
\begin{minipage}[b]{\textwidth}
\centering
%
\end{minipage}
\end{table}
\begin{table}[H]
\caption{Fichier \'{e}valu\'{e} : \tiny{\url{equipe_3_tache_2_execution_3-testing.sh-title-only-ref-silence-bandstop-aggr-eucl.prexml.xml}}}
\label{tab:equipe_3_tache_2_execution_3-testing.sh-title-only-ref-silence-bandstop-aggr-eucl.prexml.xml}
\small
\begin{minipage}[b]{\textwidth}
\centering
%
\end{minipage}
\end{table}
\begin{table}[H]
\caption{Fichier \'{e}valu\'{e} : \tiny{\url{equipe_3_tache_2_execution_3-testing.sh-title-only-ref-silence-bandstop-fft-cos.prexml.xml}}}
\label{tab:equipe_3_tache_2_execution_3-testing.sh-title-only-ref-silence-bandstop-fft-cos.prexml.xml}
\small
\begin{minipage}[b]{\textwidth}
\centering
%
\end{minipage}
\end{table}
\begin{table}[H]
\caption{Fichier \'{e}valu\'{e} : \tiny{\url{equipe_3_tache_2_execution_3-testing.sh-title-only-ref-silence-bandstop-fft-eucl.prexml.xml}}}
\label{tab:equipe_3_tache_2_execution_3-testing.sh-title-only-ref-silence-bandstop-fft-eucl.prexml.xml}
\small
\begin{minipage}[b]{\textwidth}
\centering
%
\end{minipage}
\end{table}
\begin{table}[H]
\caption{Fichier \'{e}valu\'{e} : \tiny{\url{equipe_3_tache_2_execution_3-testing.sh-title-only-ref-silence-low-aggr-cheb.prexml.xml}}}
\label{tab:equipe_3_tache_2_execution_3-testing.sh-title-only-ref-silence-low-aggr-cheb.prexml.xml}
\small
\begin{minipage}[b]{\textwidth}
\centering
%
\end{minipage}
\end{table}
\begin{table}[H]
\caption{Fichier \'{e}valu\'{e} : \tiny{\url{equipe_3_tache_2_execution_3-testing.sh-title-only-ref-silence-low-aggr-eucl.prexml.xml}}}
\label{tab:equipe_3_tache_2_execution_3-testing.sh-title-only-ref-silence-low-aggr-eucl.prexml.xml}
\small
\begin{minipage}[b]{\textwidth}
\centering
%
\end{minipage}
\end{table}
\begin{table}[H]
\caption{Fichier \'{e}valu\'{e} : \tiny{\url{equipe_3_tache_2_execution_3-testing.sh-title-only-ref-silence-low-fft-cheb.prexml.xml}}}
\label{tab:equipe_3_tache_2_execution_3-testing.sh-title-only-ref-silence-low-fft-cheb.prexml.xml}
\small
\begin{minipage}[b]{\textwidth}
\centering
%
\end{minipage}
\end{table}
\begin{table}[H]
\caption{Fichier \'{e}valu\'{e} : \tiny{\url{equipe_3_tache_2_execution_3-testing.sh-title-only-ref-silence-low-fft-eucl.prexml.xml}}}
\label{tab:equipe_3_tache_2_execution_3-testing.sh-title-only-ref-silence-low-fft-eucl.prexml.xml}
\small
\begin{minipage}[b]{\textwidth}
\centering
%
\end{minipage}
\end{table}
\begin{table}[H]
\caption{Fichier \'{e}valu\'{e} : \tiny{\url{equipe_3_tache_2_execution_3-testing.sh-title-only-ref-silence-noise-bandstop-aggr-eucl.prexml.xml}}}
\label{tab:equipe_3_tache_2_execution_3-testing.sh-title-only-ref-silence-noise-bandstop-aggr-eucl.prexml.xml}
\small
\begin{minipage}[b]{\textwidth}
\centering
%
\end{minipage}
\end{table}
\begin{table}[H]
\caption{Fichier \'{e}valu\'{e} : \tiny{\url{equipe_3_tache_2_execution_3-testing.sh-title-only-ref-silence-noise-bandstop-fft-eucl.prexml.xml}}}
\label{tab:equipe_3_tache_2_execution_3-testing.sh-title-only-ref-silence-noise-bandstop-fft-eucl.prexml.xml}
\small
\begin{minipage}[b]{\textwidth}
\centering
%
\end{minipage}
\end{table}
\begin{table}[H]
\caption{Fichier \'{e}valu\'{e} : \tiny{\url{equipe_3_tache_2_execution_3-testing.sh-title-only-ref-silence-noise-endp-aggr-eucl.prexml.xml}}}
\label{tab:equipe_3_tache_2_execution_3-testing.sh-title-only-ref-silence-noise-endp-aggr-eucl.prexml.xml}
\small
\begin{minipage}[b]{\textwidth}
\centering
%
\end{minipage}
\end{table}
\begin{table}[H]
\caption{Fichier \'{e}valu\'{e} : \tiny{\url{equipe_3_tache_2_execution_3-testing.sh-title-only-ref-silence-noise-endp-fft-eucl.prexml.xml}}}
\label{tab:equipe_3_tache_2_execution_3-testing.sh-title-only-ref-silence-noise-endp-fft-eucl.prexml.xml}
\small
\begin{minipage}[b]{\textwidth}
\centering
%
\end{minipage}
\end{table}
\begin{table}[H]
\caption{Fichier \'{e}valu\'{e} : \tiny{\url{equipe_3_tache_2_execution_3-testing.sh-title-only-ref-silence-noise-low-aggr-eucl.prexml.xml}}}
\label{tab:equipe_3_tache_2_execution_3-testing.sh-title-only-ref-silence-noise-low-aggr-eucl.prexml.xml}
\small
\begin{minipage}[b]{\textwidth}
\centering
%
\end{minipage}
\end{table}
\begin{table}[H]
\caption{Fichier \'{e}valu\'{e} : \tiny{\url{equipe_3_tache_2_execution_3-testing.sh-title-only-ref-silence-noise-low-fft-eucl.prexml.xml}}}
\label{tab:equipe_3_tache_2_execution_3-testing.sh-title-only-ref-silence-noise-low-fft-eucl.prexml.xml}
\small
\begin{minipage}[b]{\textwidth}
\centering
%
\end{minipage}
\end{table}
\begin{table}[H]
\caption{Fichier \'{e}valu\'{e} : \tiny{\url{equipe_3_tache_2_execution_3-testing.sh-title-only-ref-silence-noise-norm-aggr-cheb.prexml.xml}}}
\label{tab:equipe_3_tache_2_execution_3-testing.sh-title-only-ref-silence-noise-norm-aggr-cheb.prexml.xml}
\small
\begin{minipage}[b]{\textwidth}
\centering
%
\end{minipage}
\end{table}
\begin{table}[H]
\caption{Fichier \'{e}valu\'{e} : \tiny{\url{equipe_3_tache_2_execution_3-testing.sh-title-only-ref-silence-noise-norm-aggr-eucl.prexml.xml}}}
\label{tab:equipe_3_tache_2_execution_3-testing.sh-title-only-ref-silence-noise-norm-aggr-eucl.prexml.xml}
\small
\begin{minipage}[b]{\textwidth}
\centering
%
\end{minipage}
\end{table}
\begin{table}[H]
\caption{Fichier \'{e}valu\'{e} : \tiny{\url{equipe_3_tache_2_execution_3-testing.sh-title-only-ref-silence-noise-norm-fft-cheb.prexml.xml}}}
\label{tab:equipe_3_tache_2_execution_3-testing.sh-title-only-ref-silence-noise-norm-fft-cheb.prexml.xml}
\small
\begin{minipage}[b]{\textwidth}
\centering
%
\end{minipage}
\end{table}
\begin{table}[H]
\caption{Fichier \'{e}valu\'{e} : \tiny{\url{equipe_3_tache_2_execution_3-testing.sh-title-only-ref-silence-noise-norm-fft-eucl.prexml.xml}}}
\label{tab:equipe_3_tache_2_execution_3-testing.sh-title-only-ref-silence-noise-norm-fft-eucl.prexml.xml}
\small
\begin{minipage}[b]{\textwidth}
\centering
%
\end{minipage}
\end{table}
\begin{table}[H]
\caption{Fichier \'{e}valu\'{e} : \tiny{\url{equipe_3_tache_2_execution_3-testing.sh-title-only-ref-silence-noise-raw-aggr-eucl.prexml.xml}}}
\label{tab:equipe_3_tache_2_execution_3-testing.sh-title-only-ref-silence-noise-raw-aggr-eucl.prexml.xml}
\small
\begin{minipage}[b]{\textwidth}
\centering
%
\end{minipage}
\end{table}
\begin{table}[H]
\caption{Fichier \'{e}valu\'{e} : \tiny{\url{equipe_3_tache_2_execution_3-testing.sh-title-only-ref-silence-noise-raw-fft-eucl.prexml.xml}}}
\label{tab:equipe_3_tache_2_execution_3-testing.sh-title-only-ref-silence-noise-raw-fft-eucl.prexml.xml}
\small
\begin{minipage}[b]{\textwidth}
\centering
%
\end{minipage}
\end{table}
\begin{table}[H]
\caption{Fichier \'{e}valu\'{e} : \tiny{\url{equipe_3_tache_2_execution_3-testing.sh-title-only-ref-silence-norm-aggr-cos.prexml.xml}}}
\label{tab:equipe_3_tache_2_execution_3-testing.sh-title-only-ref-silence-norm-aggr-cos.prexml.xml}
\small
\begin{minipage}[b]{\textwidth}
\centering
%
\end{minipage}
\end{table}
\begin{table}[H]
\caption{Fichier \'{e}valu\'{e} : \tiny{\url{equipe_3_tache_2_execution_3-testing.sh-title-only-ref-silence-norm-aggr-eucl.prexml.xml}}}
\label{tab:equipe_3_tache_2_execution_3-testing.sh-title-only-ref-silence-norm-aggr-eucl.prexml.xml}
\small
\begin{minipage}[b]{\textwidth}
\centering
%
\end{minipage}
\end{table}
\begin{table}[H]
\caption{Fichier \'{e}valu\'{e} : \tiny{\url{equipe_3_tache_2_execution_3-testing.sh-title-only-ref-silence-norm-fft-cos.prexml.xml}}}
\label{tab:equipe_3_tache_2_execution_3-testing.sh-title-only-ref-silence-norm-fft-cos.prexml.xml}
\small
\begin{minipage}[b]{\textwidth}
\centering
%
\end{minipage}
\end{table}
\begin{table}[H]
\caption{Fichier \'{e}valu\'{e} : \tiny{\url{equipe_3_tache_2_execution_3-testing.sh-title-only-ref-silence-norm-fft-eucl.prexml.xml}}}
\label{tab:equipe_3_tache_2_execution_3-testing.sh-title-only-ref-silence-norm-fft-eucl.prexml.xml}
\small
\begin{minipage}[b]{\textwidth}
\centering
%
\end{minipage}
\end{table}
\begin{table}[H]
\caption{Fichier \'{e}valu\'{e} : \tiny{\url{equipe_3_tache_2_execution_3-testing.sh-title-only-ref-silence-raw-aggr-eucl.prexml.xml}}}
\label{tab:equipe_3_tache_2_execution_3-testing.sh-title-only-ref-silence-raw-aggr-eucl.prexml.xml}
\small
\begin{minipage}[b]{\textwidth}
\centering
%
\end{minipage}
\end{table}
\begin{table}[H]
\caption{Fichier \'{e}valu\'{e} : \tiny{\url{equipe_3_tache_2_execution_3-testing.sh-title-only-ref-silence-raw-fft-eucl.prexml.xml}}}
\label{tab:equipe_3_tache_2_execution_3-testing.sh-title-only-ref-silence-raw-fft-eucl.prexml.xml}
\small
\begin{minipage}[b]{\textwidth}
\centering
%
\end{minipage}
\end{table}

\begin{table}[H]
\caption{Consolidated results (piste2-result1-marf-title-only-journal-eval), Part 1.}
\label{tab:piste2-result1-marf-title-only-journal-eval-results1}
\small
\begin{minipage}[b]{\textwidth}
\centering
%
\end{minipage}
\end{table}

\begin{table}[H]
\caption{Consolidated results (piste2-result1-marf-title-only-journal-eval), Part 2.}
\label{tab:piste2-result1-marf-title-only-journal-eval-results2}
\small
\begin{minipage}[b]{\textwidth}
\centering
%
\end{minipage}
\end{table}

\begin{table}[H]
\caption{Consolidated results (piste2-result1-marf-title-only-journal-eval), Part 3.}
\label{tab:piste2-result1-marf-title-only-journal-eval-results3}
\small
\begin{minipage}[b]{\textwidth}
\centering
%
\end{minipage}
\end{table}

\begin{table}[H]
\caption{Fichier \'{e}valu\'{e} : \tiny{\url{equipe_3_tache_2_execution_3-testing.sh-title-only-journal-ref-bandstop-aggr-cheb.prexml.xml}}}
\label{tab:equipe_3_tache_2_execution_3-testing.sh-title-only-journal-ref-bandstop-aggr-cheb.prexml.xml}
\small
\begin{minipage}[b]{\textwidth}
\centering
%
\end{minipage}
\end{table}
\begin{table}[H]
\caption{Fichier \'{e}valu\'{e} : \tiny{\url{equipe_3_tache_2_execution_3-testing.sh-title-only-journal-ref-bandstop-aggr-eucl.prexml.xml}}}
\label{tab:equipe_3_tache_2_execution_3-testing.sh-title-only-journal-ref-bandstop-aggr-eucl.prexml.xml}
\small
\begin{minipage}[b]{\textwidth}
\centering
%
\end{minipage}
\end{table}
\begin{table}[H]
\caption{Fichier \'{e}valu\'{e} : \tiny{\url{equipe_3_tache_2_execution_3-testing.sh-title-only-journal-ref-bandstop-aggr-mink.prexml.xml}}}
\label{tab:equipe_3_tache_2_execution_3-testing.sh-title-only-journal-ref-bandstop-aggr-mink.prexml.xml}
\small
\begin{minipage}[b]{\textwidth}
\centering
%
\end{minipage}
\end{table}
\begin{table}[H]
\caption{Fichier \'{e}valu\'{e} : \tiny{\url{equipe_3_tache_2_execution_3-testing.sh-title-only-journal-ref-bandstop-fft-cheb.prexml.xml}}}
\label{tab:equipe_3_tache_2_execution_3-testing.sh-title-only-journal-ref-bandstop-fft-cheb.prexml.xml}
\small
\begin{minipage}[b]{\textwidth}
\centering
%
\end{minipage}
\end{table}
\begin{table}[H]
\caption{Fichier \'{e}valu\'{e} : \tiny{\url{equipe_3_tache_2_execution_3-testing.sh-title-only-journal-ref-bandstop-fft-eucl.prexml.xml}}}
\label{tab:equipe_3_tache_2_execution_3-testing.sh-title-only-journal-ref-bandstop-fft-eucl.prexml.xml}
\small
\begin{minipage}[b]{\textwidth}
\centering
%
\end{minipage}
\end{table}
\begin{table}[H]
\caption{Fichier \'{e}valu\'{e} : \tiny{\url{equipe_3_tache_2_execution_3-testing.sh-title-only-journal-ref-bandstop-fft-mink.prexml.xml}}}
\label{tab:equipe_3_tache_2_execution_3-testing.sh-title-only-journal-ref-bandstop-fft-mink.prexml.xml}
\small
\begin{minipage}[b]{\textwidth}
\centering
%
\end{minipage}
\end{table}
\begin{table}[H]
\caption{Fichier \'{e}valu\'{e} : \tiny{\url{equipe_3_tache_2_execution_3-testing.sh-title-only-journal-ref-low-aggr-cheb.prexml.xml}}}
\label{tab:equipe_3_tache_2_execution_3-testing.sh-title-only-journal-ref-low-aggr-cheb.prexml.xml}
\small
\begin{minipage}[b]{\textwidth}
\centering
%
\end{minipage}
\end{table}

\subsubsection{Results for training data for title-only processing}
\index{Results!training data for title-only processing}

\begin{table}[H]
\caption{Consolidated results (piste2-result1-marf-title-only-stats), Part 1.}
\label{tab:piste2-result1-marf-title-only-stats-results1}
\small
\begin{minipage}[b]{\textwidth}
\centering
%
\end{minipage}
\end{table}


\addcontentsline{toc}{section}{References}
\bibliographystyle{plain}
\bibliography{marf-deft-complete-results}

\addcontentsline{toc}{section}{Index}
\printindex

\end{document}